\documentclass[letterpaper]{article} 
\usepackage{aaai2027}  
\usepackage[hyphens]{url}  
\usepackage{graphicx} 
\usepackage{natbib}  
\usepackage{caption} 
\usepackage{booktabs}

\usepackage{amsmath}
\DeclareMathOperator*{\argmin}{arg\,min}
\DeclareMathOperator*{\argmax}{arg\,max}
\usepackage{amssymb}
\usepackage{bm}
\usepackage{multirow}
\usepackage{tabularx}
\usepackage{array}
\usepackage{enumitem}
\usepackage{microtype}

\usepackage{pdfpages}

\microtypesetup{expansion=false}

\newcommand{\method}{\textsc{FEAST}}

\title{\textsc{FEAST}: Federated Shared-Space Training for Resource-Heterogeneous Clients}

\author{
    Bostan Khan,
    Masoud Daneshtalab
}
\affiliations{
    Mälardalen University, Västerås, Sweden
}

\begin{document}

\maketitle

\begin{abstract}
Federated learning (FL) must serve devices with varying computational capabilities. A fixed model cannot suit all devices, while training one model per deployment limit is costly. Federated supernet training instead learns one elastic model with differently sized subnetworks, then deploys a suitable one to each device. When client inference budgets differ, however, parameters exclusive to high-cost subnetworks are reachable by fewer clients. We propose \method{}, a federated shared-space training framework that counters this imbalance by jointly training multiple subnetworks within each client's limit. Budget-tailored sub-supernet routing sends only the relevant supernet portion, and sparse aggregation merges the returned parameter slices. The trained supernet directly serves the subnetworks used during federation and supports post-hoc extraction of additional subnetworks without federated retraining. We further show that independently assigning clients' training-data volumes and inference budgets can distort accuracy--inference-cost comparisons in heterogeneous FL simulations, and introduce a one-parameter $\gamma$-allocation protocol to control this coupling. In our experimental setup, the SuperFedNAS and DeepFedNAS supernet training procedures remain near chance at 25M and reach at most $17.09\%$ at $596$M inference MACs; \method{} reaches $71.06\%$ at $596$M, $2.4$ points above the strongest model-heterogeneous weight-sharing baseline at its largest tier. Across CIFAR-100, CINIC-10, and TinyImageNet-200, \method{} achieves the highest population-averaged accuracy among the evaluated weight-sharing methods when each client receives its largest affordable subnetwork. Sub-supernet routing reduces aggregate model-parameter traffic by $6.8\times$ relative to full-supernet transmission.
\end{abstract}


\section{Introduction}
\label{sec:introduction}

Federated learning (FL) trains a shared model from decentralized data without
moving raw examples to a central server~\cite{mcmahan2017fedavg,kairouz2021advances}.
Yet one architecture is a poor fit for devices with different computational
capabilities~\cite{kairouz2021advances,li2020federated}: a model small enough
for constrained devices underuses more capable ones, whereas a large model may
be infeasible for much of the population. Training a separate federation for
each deployment limit is costly. Federated supernet training offers a
train-once, deploy-many alternative: learn one elastic network containing
parameter-sharing subnetworks of different sizes, then serve a suitable
subnetwork to each device.

Prior federated-supernet methods, including
SuperFedNAS~\cite{khare2024superfednas} and
DeepFedNAS~\cite{khan2026deepfednas}, pursue this train-once, deploy-many
objective. Client-specific deployment limits, however, create an optimization
asymmetry: parameters shared by low-cost subnetworks are reachable by nearly
all clients, whereas parameters exclusive to high-cost subnetworks are
reachable only by the high-budget minority. The amount and distribution of
data held by that minority further determine the effective training signal
available to those parameters. Our central question is therefore how to train
a directly served elastic model when client budgets and data allocation
jointly produce highly unequal parameter exposure.

We formulate budget-constrained federated supernet training in terms of one
elastic parameter tensor, overlapping subnetworks, client-specific affordable
sets, and budget-dependent parameter reachability; we refer to this setting as
a \emph{federated shared model space} (FSMS). To train it, we introduce
\method{} (FEderAted Shared-space Training). On each mini-batch, a client
jointly trains the globally smallest, an affordable random intermediate, and
its largest affordable cached variant, using the local maximum to distill the
smaller variants. Budget-tailored routing transmits only the parameter envelope
spanning the client's affordable variants, and position-wise sparse,
sample-count-weighted aggregation maps returned slices into the global tensor.
The resulting supernet directly serves the cached variants and supports
post-hoc extraction of additional variants without federated retraining.

Resource-heterogeneous simulations also assign each client a budget and
training data, commonly generating them independently even though system and
data heterogeneity can interact in practice~\cite{maeng2022fair,wang2025intertwined}.
In a low-budget-heavy population, this independence gives the high-budget
minority no larger expected data share, although only those clients can reach
parameters exclusive to high-cost variants. We therefore introduce a
one-parameter allocation family in which $\gamma=0$ recovers
budget-independent allocation and $\gamma=1$ makes expected data share
proportional to capped budget (Sec.~\ref{sec:protocol}).

\begin{figure}[t]
    \centering
    \includegraphics[width=\columnwidth]{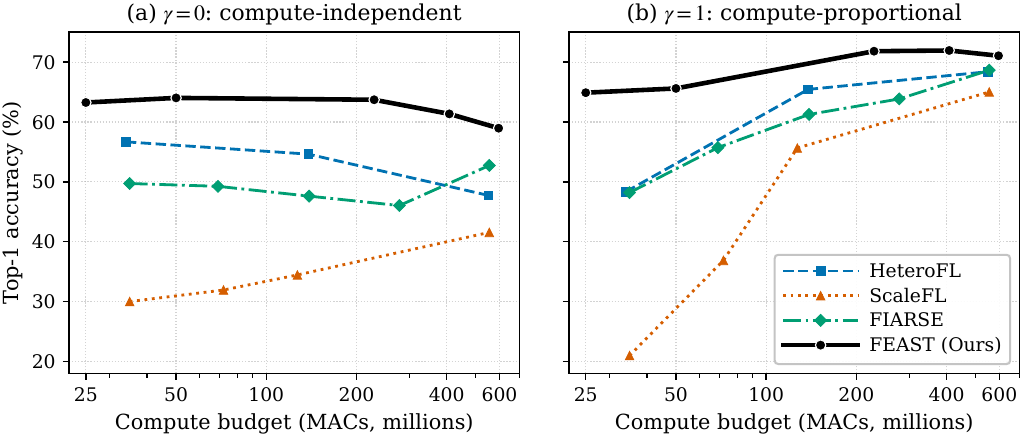}
    \caption{Effect of compute--data allocation on
    CIFAR-100~\cite{krizhevsky2009cifar} accuracy over the shared
    $\geq25$M-MAC range for \method{},
    HeteroFL~\cite{diao2021heterofl},
    ScaleFL~\cite{ilhan2023scalefl}, and FIARSE~\cite{wu2024fiarse}.
    $\gamma=0$ makes expected data volume independent of client budget,
    whereas $\gamma=1$ makes it proportional to the capped budget. Changing
    only this relationship raises accuracy at each method's largest evaluated
    point.}
    \label{fig:intro_curves}
\end{figure}

Figure~\ref{fig:intro_curves} shows that changing only this allocation
relationship materially changes the observed accuracy--inference-cost curves
of all four methods. The compute--data relationship is therefore a
consequential evaluation choice, rather than a neutral implementation detail.

Our contributions are:
\begin{itemize}[leftmargin=*,nosep]
    \item We formulate budget-constrained federated supernet training as an
    FSMS and establish nested budget reachability. We characterize expected
    routed and activation mass and derive a conditional finite-horizon
    stationarity bound whose dependence on the minimum routed computation
    across served parameters makes the cost of unequal coverage explicit.
    \item We propose \method{}, which combines local minimum--random--maximum
    co-training and in-place distillation with budget-tailored sub-supernet
    routing and sparse aggregation to improve accuracy across the directly
    served model range.
    \item We introduce a $\gamma$-controlled allocation protocol for studying
    how client inference capability and training-data allocation jointly shape
    the accuracy--inference-cost relationship.
    \item Across CIFAR-100, CINIC-10, and TinyImageNet-200, \method{} achieves
    the strongest population-averaged accuracy among the evaluated
    weight-sharing methods under largest-affordable serving. Its advantage
    persists under equal total training computation on CIFAR-100, while
    sub-supernet routing substantially reduces communication relative to
    full-supernet transmission.
\end{itemize}


\section{Related Work}
\label{sec:related_work}

\noindent\textbf{Federated NAS and federated supernets.}
Federated NAS performs architecture search over decentralized, often non-IID
client data, while resource-aware variants additionally account for device
constraints~\cite{he2020towards,yuan2022resource}. DC-NAS systematically
samples the supernet search space and applies channel pruning to reduce client
training cost~\cite{venkatesha2023dcnas}; FINCH uses hierarchical
search~\cite{liu2024finch}; and PerFedRLNAS and Peaches personalize client
architectures~\cite{yao2024perfedrlnas,yan2024peaches}. The closest
predecessors to our train-once, deploy-many setting are SuperFedNAS and
DeepFedNAS. SuperFedNAS co-trains a supernet and uses predictor-guided
post-training search for deployment targets~\cite{khare2024superfednas};
DeepFedNAS replaces random sampling with a Pareto-guided training cache and
uses its fitness function for predictor-free
selection~\cite{khan2026deepfednasjournal,khan2026deepfednas}. Resource-aware
one-shot methods already restrict each client to resource-feasible subnetworks
during federation; HAPFNAS combines this assignment with server-side
distillation and personalized search~\cite{yang2025hapfnas}. A recent
supernet-only framework reinterprets residual networks as supernets and
directly serves depth-truncated subnetworks matched to client
capability~\cite{chen2025supernetonly}. Building on these
capabilities, \method{} trains a single elastic model whose variants are served
directly under client-specific inference budgets. Because parameters used only
by larger variants are reachable by fewer clients, \method{} co-trains multiple
affordable variants within each client and routes only the parameter slice
covering the variants that client can train.

\noindent\textbf{Elastic networks and shared parameter spaces.}
Centralized elastic networks provide the underlying training mechanisms.
Universally Slimmable Networks combine the sandwich rule with in-place
distillation~\cite{Yu2019ICCV}; Once-for-All trains once and specializes to
hardware constraints~\cite{cai2020once}; and BigNAS optimizes child models for
direct deployment without fine-tuning~\cite{yu2020bignas}. Weight-sharing NAS
reduces search cost but can suffer optimization interference and unreliable
inherited-subnetwork rankings~\cite{bender2018understanding,Weight-Sharing}.
These methods assume centralized access to the full supernet. FEAST adapts
their training principles to a budget-constrained federation, where only
high-budget clients can reach parameters exclusive to large variants.

\noindent\textbf{Model-heterogeneous federated learning.}
Model-heterogeneous FL matches client workloads to local capability. HeteroFL
uses nested width-scaled networks~\cite{diao2021heterofl}; FjORD uses ordered
dropout~\cite{horvath2021fjord}; FedRolex rotates model
slices~\cite{alam2022fedrolex}; ScaleFL adapts depth and width with
self-distillation~\cite{ilhan2023scalefl}; and FIARSE performs importance-aware
extraction~\cite{wu2024fiarse}. OFA-FL jointly optimizes resource-specific
subnetworks within one federated model~\cite{varma2023ofafl}, while FedEL
progressively covers a fixed network using runtime-budgeted training
windows~\cite{zhang2025fedel}. Split-Mix instead forms larger inference models
by ensembling multiple separately parameterized base models at
inference~\cite{hong2022splitmix}, so it falls outside our weight-sharing
quantitative comparison. These methods support heterogeneous client workloads.
Within this broader line, FEAST focuses on unequal client exposure across
overlapping regions of a structured depth-width-expansion supernet while
serving variants directly from the shared tensor.

\noindent\textbf{Coupled system and data heterogeneity.}
FL research has long recognized that system and statistical heterogeneity can
interact and bias participation or
optimization~\cite{kairouz2021advances,li2020federated}. Maeng et al. study
their interdependence in federated recommendation~\cite{maeng2022fair}, while
Wang and Gao analyze intertwined device and data heterogeneity under stale
updates~\cite{wang2025intertwined}. Resource-heterogeneous evaluations
nevertheless commonly specify compute tiers and data partitions separately. To
our knowledge, no prior resource-heterogeneous FL evaluation introduces an
explicit parameter controlling the dependence between expected client data
volume and inference budget. Our $\gamma$-parameterized protocol provides this
control and is applied identically to FEAST and all evaluated baselines.


\section{Method}
\label{sec:method}

\subsection{Resource-Heterogeneous Evaluation Protocol}
\label{sec:protocol}

\paragraph{Client population and MAC budgets.}
Client $i$ has an inference budget $b_i$, defined as the maximum per-example
inference cost, measured in MACs, of any individual variant it may train or
deploy. This per-variant ceiling does not bound cumulative local-pass cost, as
discussed in Sec.~\ref{sec:cotraining}. To model a
controlled low-end-heavy population, we draw $N$ budget ranks from
Zipf($s=1.2$) and map them to $[25\mathrm{M},1500\mathrm{M}]$ MACs; Pisces
uses the same exponent for moderately skewed client latency~\cite{jiang2022pisces}.
Served variants span approximately 25M--600M MACs. Raw $b_i$ determines
affordability, while allocation uses
$b_i^{\mathrm{alloc}}=\min(b_i,B_{\mathrm{cap}})$ with
$B_{\mathrm{cap}}=600$M, preventing capability beyond the served range from
increasing allocation weight. Supplementary Sec.~A gives the exact
rank-to-budget mapping.

\vspace{-4pt}
 
\paragraph{\(\gamma\)-allocation.}
Standard simulations often construct client data partitions independently of
client budgets. In a low-end-heavy population, this gives the smaller subset of
clients that can train high-cost variants no larger expected data share, which
can limit optimization of variant-exclusive parameters. We control the
dependence between a client's expected data volume and budget through
\begin{equation}
    q_i(\gamma)
    =
    \frac{\left(b_i^{\mathrm{alloc}}\right)^{\gamma}}
    {\sum_{j=1}^{N}\left(b_j^{\mathrm{alloc}}\right)^{\gamma}},
    \qquad
    b_i^{\mathrm{alloc}}=\min(b_i,B_{\mathrm{cap}}),
    \label{eq:gamma_alloc}
\end{equation}
where $\gamma\geq0$. Setting $\gamma=0$ yields uniform base weights
$q_i(0)=1/N$, while $\gamma=1$ makes the initial expected data share proportional
to the capped budget. We use $\gamma=1$ as a conservative first-order default,
not as a universal deployment law, and apply the same allocation protocol to
every evaluated method.

\vspace{-4pt}
 
\paragraph{Budget-aware Dirichlet partition.}
Let $N_{\mathrm{train}}$ denote the number of examples in the client-training
pool after excluding server validation and BN-calibration data, and let class
$k$ contain $N_k$ examples. We draw its initial client proportions as
\begin{equation}
    \widetilde{\mathbf{p}}_k
    \sim
    \mathrm{Dir}\!\left(c_{k,1},\ldots,c_{k,N}\right),
    \qquad
    c_{k,i}=\alpha_{\mathrm{data}}N_kq_i(\gamma).
    \label{eq:asym_dirichlet}
\end{equation}
Before integer assignment,
$\mathbb{E}[\widetilde p_{k,i}]=q_i(\gamma)$ and
$\sum_i c_{k,i}=\alpha_{\mathrm{data}}N_k$. Thus, $\gamma$ changes the target
mean vector while preserving total per-class precision. It also changes
component variances and hence the realized variability of shard sizes and
per-class client allocations. Because local optimizer-step counts and
sample-weighted aggregation depend on realized shard size, comparisons across
$\gamma$ characterize the complete coupled allocation protocol rather than a
data-quantity-only intervention. Integer assignment uses
$q_i(\gamma)N_{\mathrm{train}}$ as a per-client balancing cap while producing
disjoint shards, so $\mathbf q(\gamma)$ is a pre-integer volume target rather
than the exact realized sample-share vector. For a balanced $K$-class
dataset, we report
$\alpha_{\mathrm{eff}}=\alpha_{\mathrm{data}}N_{\mathrm{train}}/(KN)$ and set
$\alpha_{\mathrm{data}}$ so that $\alpha_{\mathrm{eff}}=0.45$ on all three
datasets. Supplementary Sec.~A gives the complete assignment and balancing
procedure.

\subsection{Federated Shared Model Space}
\label{sec:fsms}

We formulate federated supernet training in terms of a shared elastic parameter
tensor, a family of overlapping inference-budget-matched variants activated
from that tensor, and a federation rule that restricts each client to affordable
variants and maps its returned parameter values to the corresponding global
positions. We refer to this formulation as a federated shared model space
(FSMS).

\vspace{-4pt}
 
\paragraph{Shared elastic family and fixed cache.}
The server maintains an OFA-style ResNet
$\mathcal{M}(\bm{\theta})$ with $|\bm{\theta}|=55.68$M parameters. An
architecture $\mathbf{a}=(\mathbf{d},\mathbf{e},\mathbf{w})$ selects stage
depths, expansion ratios, and widths, activating a shared slice
$\bm{\theta}_{\mathbf{a}}\subseteq\bm{\theta}$; variants therefore maintain
no private weights. As in elastic-network training~\cite{cai2020once,Yu2019ICCV},
batch-normalization statistics are recalibrated separately for each evaluated
operating point. The four-stage supernet ($S=4$) uses base channels
$[128,256,512,1024]$, additional-block counts
$d_\ell\in\{0,\ldots,8\}$, stage-level expansion ratios
$e_\ell\in\{0.10,0.14,0.18,0.22,0.25\}$, and width multipliers
$\{0.1,0.2,\ldots,1.0\}$. Let $\mathcal H$ denote the resulting architecture
space. From $\mathcal H$, \method{} uses a fixed cache
$\mathcal A\subset\mathcal H$ of 56 variants spanning approximately
25M--600M MACs, generated once before federation by retargeting the
Pareto-guided DeepFedNAS procedure~\cite{khan2026deepfednas,khan2026deepfednasjournal}.
The cache fixes local choices, routing, and the reported serving and evaluation
points; no architecture search occurs during federation. It does not delimit
deployment: additional subnetworks can be extracted from the trained tensor
and BN-recalibrated without further federated training, although we evaluate
only $\mathcal A$. Supplementary Sec.~B gives the search space, offline
objective, and cache construction.

\vspace{-4pt}
 
\paragraph{Affordable variants and structural asymmetry.}
Client $i$ may activate only
\begin{equation}
    \mathcal{A}_i
    =
    \left\{
        \mathbf{a}\in\mathcal{A}
        \;\middle|\;
        \mathrm{MAC}(\mathbf{a})\leq b_i
    \right\},
    \label{eq:affordable}
\end{equation}
which we call its affordable variant set. In every realized population used
here, $\mathcal A_i$ is nonempty and contains the globally smallest cached
variant. Affordability uses the raw client budget $b_i$, not the
allocation-capped quantity $b_i^{\mathrm{alloc}}$ in
Eq.~\eqref{eq:gamma_alloc}; the cap affects data allocation only. A parameter
position is reachable by client $i$ when some variant in $\mathcal A_i$
activates it.
Consequently, low-cost regions are shared broadly, whereas high-cost-exclusive
regions depend on the high-budget minority. \method{} addresses this structural
asymmetry through budget-tailored routing and multi-variant co-training.

\begin{figure*}[t]
    \centering
    \includegraphics[width=\textwidth]{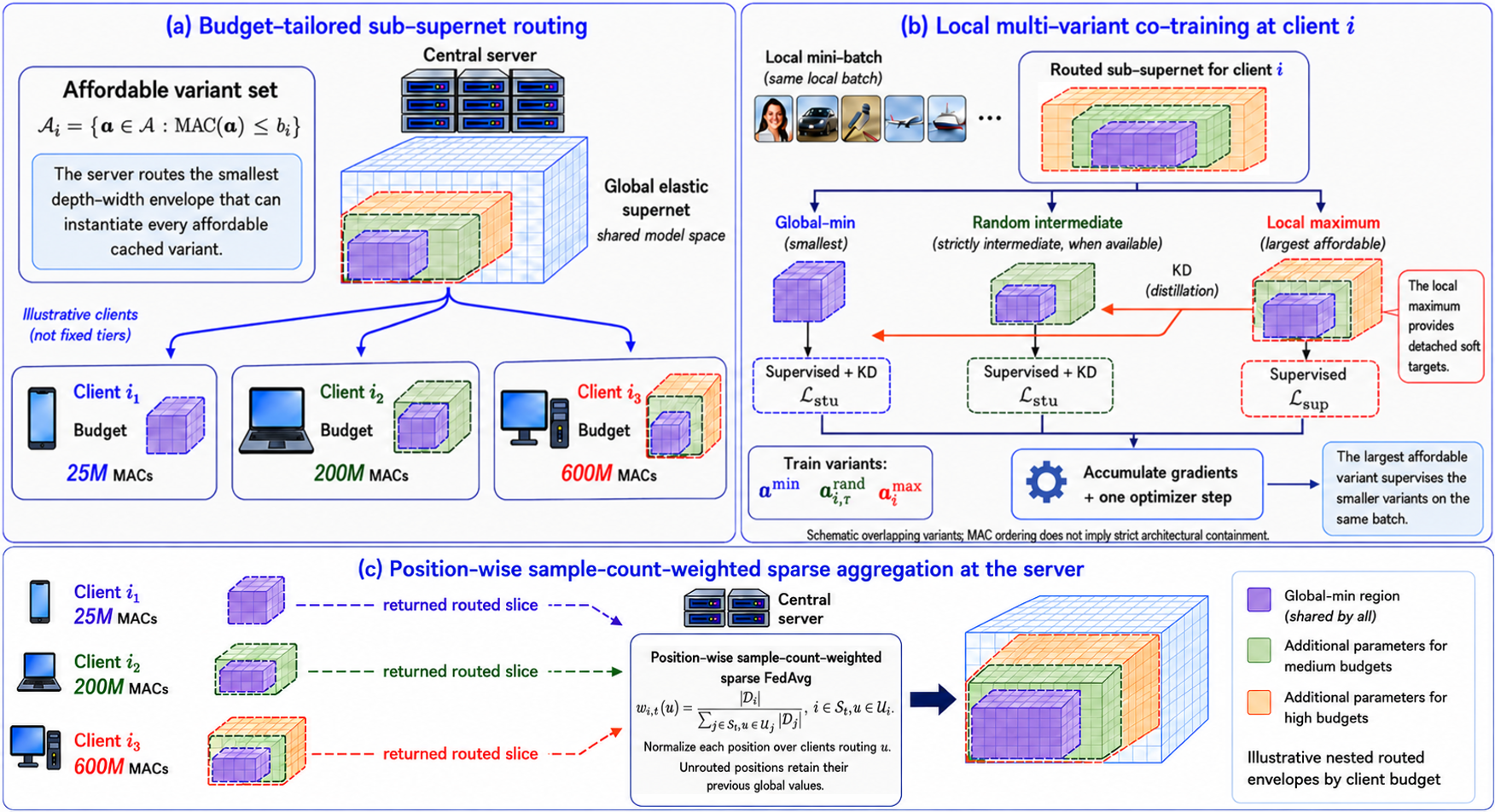}
    \caption{One \method{} round. (a) The server routes each selected client a
    depth-width-bounded sub-supernet covering its affordable variants. (b) On
    each mini-batch, the client trains the global minimum, its local maximum,
    and an affordable random intermediate when available; the local maximum
    supplies detached soft targets to the smaller variants. (c) Returned
    slices are mapped to global positions and combined by sample-count-weighted
    sparse FedAvg.}
    \label{fig:feast_method}
    \vspace{-8pt}
\end{figure*}

Figure~\ref{fig:feast_method} summarizes one round. In the following, $t$
indexes federation rounds and $\tau$ indexes local mini-batches.

We next describe budget-tailored routing and sparse aggregation
(Sec.~\ref{sec:communication}), followed by local multi-variant co-training
(Sec.~\ref{sec:cotraining}).

\subsection{FEAST: Budget-Tailored Sub-Supernet Routing}
\label{sec:communication}

\paragraph{Client-specific routing.}
For selected client $i$, \method{} derives coordinate-wise depth and width
bounds over its affordable set $\mathcal{A}_i$:
\begin{equation}
\begin{aligned}
    d^{*}_{i,\ell}
    &=
    \max_{\mathbf{a}\in\mathcal{A}_i} d_\ell(\mathbf{a}),
    && \ell=1,\ldots,S,
    \\
    w^{*}_{i,j}
    &=
    \max_{\mathbf{a}\in\mathcal{A}_i} w_j(\mathbf{a}),
    && j=0,\ldots,S.
\end{aligned}
\label{eq:sub_supernet_bounds}
\end{equation}
These maxima define the tightest depth--width envelope, under the implemented
parameterization, that contains every affordable cached variant. They may arise
from different variants, so the envelope can represent uncached combinations,
but local sampling remains restricted to $\mathcal{A}_i$. The server
instantiates the reduced supernet, sizes retained tensors to support all
globally allowed stage-level expansion ratios, copies the corresponding leading
slices of the global weights, and records a map $\Omega_i$ from local to global
positions.
We denote the covered global positions by $\mathcal{U}_i$. If the bounds are
global, routing reduces to the full supernet. Exact tensor construction is
provided in Supplementary Sec.~C.

\vspace{-4pt}
 
\paragraph{Nested routed support.}
For global position $u$, let
\begin{equation}
    r_i(u)=\mathbb{I}\{u\in\mathcal U_i\}
    \label{eq:method_routed_support}
\end{equation}
indicate routing eligibility.

\noindent\textbf{Proposition 1 (Nested budget reachability).}
Because affordability is thresholded by budget and routed tensors are nested
leading slices, $b_i\leq b_j$ implies
\begin{equation}
    \mathcal A_i\subseteq\mathcal A_j,
    \qquad
    \mathcal U_i\subseteq\mathcal U_j,
    \qquad
    r_i(u)\leq r_j(u)\quad\forall u.
    \label{eq:method_nested_support}
\end{equation}
Hence, positions exclusive to larger variants are routed only by an upper tail
of the budget-ordered population. Supplementary Sec.~F provides the proof and
finite-population coverage analysis.

\vspace{-4pt}
 
\paragraph{Sparse sample-weighted aggregation.}
Let $\mathcal D_i$ denote client $i$'s realized local shard and
$n_i=|\mathcal D_i|$. In round $t$, the server samples a cohort
$\mathcal S_t$ of $m$ clients uniformly without replacement from the $N$
clients. Let $\bar{\bm{\theta}}'_i[u]$ be client $i$'s returned value mapped
to global position $u$. The server averages each position only over selected
clients whose routed envelopes contain it:
\begin{equation}
    \bm{\theta}^{t+1}[u]
    =
    \begin{cases}
    \displaystyle
    \frac{
        \sum_{\substack{i\in\mathcal{S}_t\\u\in\mathcal{U}_i}}
        n_i\,\bar{\bm{\theta}}'_i[u]
    }{
        \sum_{\substack{i\in\mathcal{S}_t\\u\in\mathcal{U}_i}}
        n_i
    },
    & \exists i\in\mathcal{S}_t:\,u\in\mathcal{U}_i,
    \\[3mm]
    \bm{\theta}^{t}[u],
    & \text{otherwise}.
    \end{cases}
    \label{eq:sparse_update}
\end{equation}
Normalization is therefore position-specific and sample-count-weighted;
positions not routed in the round retain their previous global values.
Aggregation follows routed coverage rather than local activation. Thus, a
routed but unactivated position is still included through its returned tensor
value. Supplementary Secs.~C and~F give the accumulator implementation and
routed-versus-activated analysis.

\paragraph{Communication accounting.}
Ignoring metadata and using the same parameter precision in both directions,
routing changes the full-supernet round-trip payload $2mP$ to
$2\sum_{i\in\mathcal S_t}P_i$, where client $i$ receives $P_i$ of $P$
parameters. Section~\ref{sec:communication_results} reports the reductions;
Supplementary Sec.~C gives the derivation.

\subsection{FEAST: Local Multi-Variant Co-Training}
\label{sec:cotraining}

\paragraph{Budget-conditioned variant selection.}
Figure~\ref{fig:feast_method}(b) illustrates the local rule. On each
mini-batch, client $i$ trains the globally smallest cached variant, its
largest affordable variant, and one affordable random intermediate when one
exists. The two endpoints are
\begin{align}
    \mathbf{a}^{\min}
    &=
    \argmin_{\mathbf{a}\in\mathcal{A}}
    \mathrm{MAC}(\mathbf{a}),
    \notag\\
    \mathbf{a}^{\max}_i
    &=
    \argmax_{\mathbf{a}\in\mathcal{A}_i}
    \mathrm{MAC}(\mathbf{a}).
    \label{eq:variant_endpoints}
\end{align}
Because $\mathbf{a}^{\min}\in\mathcal A_i$ for every realized client, all
selected clients optimize the same low-cost endpoint, while
$\mathbf{a}^{\max}_i$ directly optimizes the highest-cost endpoint permitted by
$b_i$. Let
$\mathcal{A}^{\mathrm{int}}_i$ contain the cached variants whose MAC costs lie
strictly between these endpoints. If this set is nonempty, the client samples
$\mathbf{a}^{\mathrm{rand}}_{i,\tau}$ uniformly from it for each mini-batch
$\tau$; otherwise, the intermediate pass is skipped. When the endpoints
coincide, the step reduces to supervised training of $\mathbf{a}^{\min}$.

\vspace{-4pt}
 
\paragraph{Local objective and distillation.}
The same mixed-label tuple $(\tilde{x},y_a,y_b,\lambda_{\mathrm{mix}})$,
produced by the mixup/CutMix augmentation described in
Sec.~\ref{sec:exp_setup}, is reused across all applicable passes. Let
$z_{i,\tau}(\mathbf a)=f(\tilde{x};\bm{\theta}_i,\mathbf a)$; for logits
$z$, define
\begin{align}
    \mathcal{L}_{\mathrm{sup}}(z)
    &=
    \lambda_{\mathrm{mix}}\mathcal{L}_{\mathrm{CE}}(z,y_a)
    +(1-\lambda_{\mathrm{mix}})\mathcal{L}_{\mathrm{CE}}(z,y_b),
    \notag\\
    \hat{p}^{\max}_{i,\tau}
    &=
    \mathrm{softmax}\!\left(
        \mathrm{sg}(z_{i,\tau}(\mathbf{a}^{\max}_i))
    \right),
    \notag\\
    \mathcal{L}_{\mathrm{KD}}(z,\hat p)
    &=
    -\sum_c \hat p_c\log \mathrm{softmax}(z)_c,
    \notag\\
    \mathcal{L}_{\mathrm{stu}}(z)
    &=
    (1-\rho_{\mathrm{kd}})\mathcal{L}_{\mathrm{sup}}(z)
    +\rho_{\mathrm{kd}}
    \mathcal{L}_{\mathrm{KD}}(z,\hat p^{\max}_{i,\tau}).
    \label{eq:student_loss}
\end{align}
Here $\mathrm{sg}(\cdot)$ denotes stop-gradient. The local maximum is evaluated first and optimized only with
$\mathcal{L}_{\mathrm{sup}}$; its detached distribution supervises the global
minimum and, when present, the random intermediate through
$\mathcal{L}_{\mathrm{stu}}$. We use $\rho_{\mathrm{kd}}=0.5$ and temperature
$T=1$ (no scaling). The client backpropagates each applicable loss,
accumulates the gradients, clips their combined $L_2$ norm to $10$, and takes
one optimizer step. This adapts in-place elastic-network distillation by using
each client's local maximum, rather than the globally largest variant, as the
teacher~\cite{yu2020bignas,Yu2019ICCV}. The complete casewise objective and
client procedure are given in Supplementary Sec.~D.

Although each activated variant satisfies
$\mathrm{MAC}(\mathbf a)\leq b_i$, the cumulative cost of the sequential
passes is not constrained by $b_i$. Supplementary Sec.~D gives the complete
computation model.

\vspace{-4pt}
 
\paragraph{Structural exposure and finite-horizon result.}
Let $\mathcal V$ denote the set of up to three variants trained on one local
mini-batch, and let $\nu_i$ be its distribution for client $i$. Define the
expected activation multiplicity per local optimization step
\begin{equation}
    h_i(u)=
    \mathbb E_{\mathcal V\sim\nu_i}
    \left[
        \sum_{\mathbf a\in\mathcal V}
        \mathbb I\{u\in\bm\theta_{\mathbf a}\}
        \;\middle|\;b_i
    \right].
    \label{eq:method_activation_profile}
\end{equation}
Here $r_i(u)$ controls routing and aggregation eligibility, whereas $h_i(u)$
counts expected local variant-activation multiplicity and may exceed one.
Define the routed sample mass
$M_t^{\mathrm{route}}(u)=\sum_{i\in\mathcal S_t}n_ir_i(u)$ and the expected
activation mass
$M_t^{\mathrm{exp}}(u)=\sum_{i\in\mathcal S_t}n_ih_i(u)$.

\noindent\textbf{Proposition 2 (Expected exposure and allocation).}
Conditioned on the realized budgets and local sample counts, uniform cohort
sampling gives
\begin{equation}
\begin{aligned}
    \mathbb E[M_t^{\mathrm{route}}(u)]
    &=\frac{m}{N}\sum_i n_i r_i(u),
    \\
    \mathbb E[M_t^{\mathrm{exp}}(u)]
    &=\frac{m}{N}\sum_i n_i h_i(u).
\end{aligned}
\label{eq:method_expected_exposure_compact}
\end{equation}
For the idealized pre-integer counterparts, the corresponding sums replace
$n_i$ by $N_{\mathrm{train}}q_i(\gamma)$, and the routed allocation share
$W_u^{\mathrm{route}}(\gamma)=\sum_iq_i(\gamma)r_i(u)$ is nondecreasing in
$\gamma$. This does not imply monotone activation exposure because $h_i(u)$
can change nonmonotonically with the affordable sampling pool.

\textbf{Theorem 1 (Routed-scale finite-horizon supervised stationarity).}
On the served support and before BN recalibration, let
$F_{\mathrm{sup}}$ be the sample-weighted supervised multi-variant reference
objective, excluding KD, and assume it is lower bounded by $F_{\inf}$ and
$L_F$-smooth. Let $\beta_i(u)$ be client $i$'s expected position-wise
aggregation weight and $B_i^{\mathrm{loc}}$ its local step count. Define
$\psi_u=\sum_i\beta_i(u)B_i^{\mathrm{loc}}$ and
$\bm{\Psi}=\operatorname{diag}(\psi_u)$. Let
$0<\psi_{\min}\leq\psi_{\max}$ denote the minimum and maximum of $\psi_u$ over
served positions. Set
$\mathcal H_T=\sum_{t<T}\eta_t$, $\mathcal Q_T=\sum_{t<T}\eta_t^2$, and
$\Delta_F=F_{\mathrm{sup}}(\bm\theta^0)-F_{\inf}$. Assume that the conditional
mean update equals
$-\eta_t\bm{\Psi}\nabla F_{\mathrm{sup}}(\bm\theta^t)$ plus a discrepancy whose
norm is at most $\eta_t\varepsilon$, and that the corresponding centered update
noise, after division by $\eta_t$, has second moment at most
$\sigma_{\mathrm{round}}^2$. If
$0<\eta_t\leq(4L_F\psi_{\max})^{-1}$, then
\begin{multline}
 \frac{1}{\mathcal H_T}\sum_{t<T}\eta_t
 \mathbb E\|\nabla F_{\mathrm{sup}}(\bm\theta^t)\|_2^2
 \\
 \leq\mathcal O\!\left(
 \frac{\Delta_F}{\psi_{\min}\mathcal H_T}
 +\frac{\varepsilon^2}{\psi_{\min}^2}
 +\frac{L_F\mathcal Q_T
 (\varepsilon^2+\sigma_{\mathrm{round}}^2)}
 {\psi_{\min}\mathcal H_T}
 \right).
 \label{eq:method_finite_horizon_bound}
\end{multline}
This conditional result makes the dependence on $\psi_{\min}$ explicit: a
smaller minimum routed computation scale loosens the finite-horizon guarantee.
The residual also depends on discrepancy and round noise, and the theorem does
not certify a small $\varepsilon$. For high-cost-exclusive positions,
increasing $\gamma$ shifts idealized allocation toward the high-budget clients
that route them. Because $\psi_u$ also depends on aggregation weights and local
step counts, this protocol change can affect $\psi_u$, but the theorem neither
proves monotonicity of $\psi_u$ nor predicts the numerical accuracy gap.
Empirically, the CIFAR-100 largest-budget-minus-smallest-budget accuracy gap
changes from $-4.29$ to $+6.17$ percentage points as $\gamma$ increases from
$0$ to $1$.
Supplementary Sec.~F gives exact assumptions, constants, proofs, and
realized-partition qualifications.



\section{Experiments}
\label{sec:experiments}

We test compute--data allocation, deployment accuracy across budgets and
clients, and the mechanisms underlying \method{}. Supplementary Sec.~E gives
dataset settings, post-hoc extraction, cost controls, parameter footprints,
and full sensitivity results; communication is derived in Supplementary
Sec.~C.

\subsection{Experimental Setup}
\label{sec:exp_setup}

We evaluate CIFAR-100~\cite{krizhevsky2009cifar},
CINIC-10~\cite{darlow2018cinic}, and
TinyImageNet-200~\cite{tinyimagenet2015} with 100 clients, 10 sampled per
round, one local epoch, batch size 64, and 4000 rounds. Budgets follow
Zipf($s=1.2$) over 25M--1500M MACs; allocation uses the 600M cap,
$\gamma=1$, and matched $\alpha_{\mathrm{eff}}=0.45$. All methods use
RandAugment$(2,6)$ with alternating Mixup and CutMix and matched optimization
except FIARSE, which retains its original zero-weight-decay setting. For every
method, we select the best
validation checkpoint and deterministically recalibrate BN statistics using
disjoint server-held samples. Supplementary Sec.~E gives additional dataset
and optimization details.

We compare \method{} with HeteroFL~\cite{diao2021heterofl},
ScaleFL~\cite{ilhan2023scalefl}, and FIARSE~\cite{wu2024fiarse} at measured
native operating points without interpolation: 56 cached \method{} variants,
five HeteroFL widths, four ScaleFL depth--width levels, and FIARSE's
continuous mask family. We additionally transfer
SuperFedNAS (SFN)~\cite{khare2024superfednas} and DeepFedNAS
(DFN)~\cite{khan2026deepfednas} to our expanded supernet and experimental
setting, retaining their random and Pareto-path samplers and MaxNet's
cosine-annealed, overlap-aware aggregation. MACs measure one served variant's
per-example inference cost; the ceilings do not bound cumulative training,
memory, latency, or energy. Supplementary Sec.~E gives complete adaptation
and operating-point details.

\begin{figure*}[t]
\centering
\includegraphics[width=0.74\textwidth]{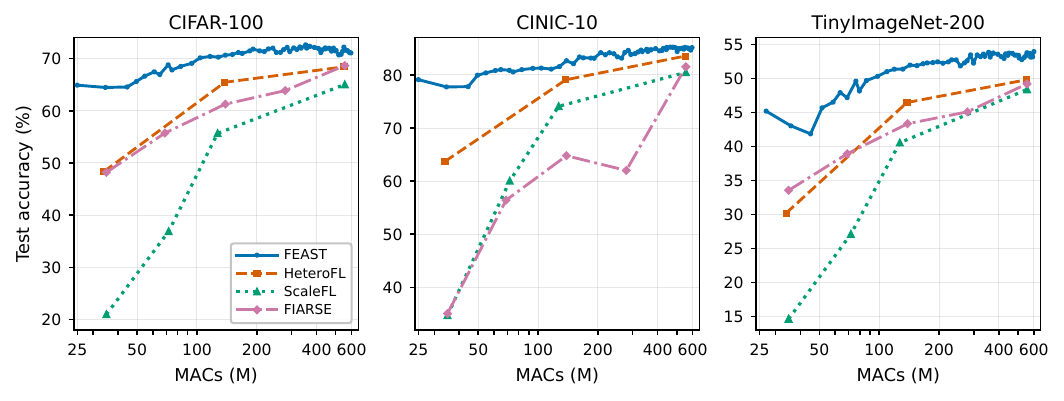}
\caption{Accuracy at every measured operating point under $\gamma=1$ for
\method{} and the three weight-sharing baselines. Lines connect measured
points only to guide the eye; sub-25M points lie outside \method{}'s studied
deployment range.}
\label{fig:main_accuracy}
\end{figure*}

\subsection{Allocation and Deployment Accuracy}
\label{sec:gamma_results}
\label{sec:main_accuracy}
\label{sec:population_results}

\paragraph{Allocation.}
Changing only $\gamma$ from 0 to 1 raises the largest-model accuracy of
\method{}, HeteroFL, ScaleFL, and FIARSE by 12.10, 20.66, 23.45, and 15.91
points, respectively (Fig.~\ref{fig:intro_curves}). For \method{}, the 596M
result progresses from 58.96\% at $\gamma=0$ to 63.79\%, 65.76\%, 71.06\%,
and 74.83\% at $\gamma\in\{0.25,0.5,1,1.5\}$, showing gradual recovery.
$\gamma=1$ is a proportional default, not the best-performing sweep value
(complete sweep in Supplementary Sec.~E).

At each method's largest evaluated point, \method{} remains ahead of the
strongest result among HeteroFL, ScaleFL, and FIARSE under both tested
allocation settings, with a 6.22-point lead at $\gamma=0$ and a 2.41-point
lead at $\gamma=1$.

\vspace{-4pt}
 
\paragraph{Budgets and clients.}
Figure~\ref{fig:main_accuracy} shows that \method{} peaks at $72.61\%$ on
CIFAR-100, $85.19\%$ on CINIC-10, and $53.94\%$ on TinyImageNet. Across the
shared $\geq25$M-MAC range, its smallest-to-largest accuracy changes are
$+6.17$, $+6.03$, and $+8.75$ percentage points, whereas the corresponding
weight-sharing baseline ranges are $+20.07$--$+44.03$,
$+19.81$--$+46.42$, and $+15.66$--$+33.74$ points, respectively.
Table~\ref{tab:population_largest} reports largest-affordable serving over
the fixed client population. For this metric, HeteroFL and ScaleFL use their
fixed native tiers, whereas FIARSE is evaluated on a dense measured MAC grid
over its continuous mask family. \method{} leads the strongest
weight-sharing baseline by $12.61$, $26.09$, and $9.44$ points on CIFAR-100,
CINIC-10, and TinyImageNet, respectively.

\begin{table}[b]
\centering
\footnotesize
\setlength{\tabcolsep}{1.8pt}
\begin{tabular*}{\columnwidth}{@{\extracolsep{\fill}}lccccc@{}}
\toprule
Dataset & HeteroFL & ScaleFL & FIARSE & \method{} &
$\Delta_{\mathrm{best}}$ \\
\midrule
CIFAR-100    & 41.50 & 35.78 & 55.13 & \textbf{67.74} &
\textbf{+12.61} \\
CINIC-10     & 54.87 & 52.16 & 53.91 & \textbf{80.96} &
\textbf{+26.09} \\
TinyImageNet & 27.38 & 25.83 & 38.74 & \textbf{48.18} &
\textbf{+9.44} \\
\bottomrule
\end{tabular*}
\caption{Population-averaged test accuracy (\%) under largest-affordable
serving for the fixed Zipf($s=1.2$) client population.
$\Delta_{\mathrm{best}}$ is the \method{} margin over the strongest
weight-sharing baseline.}
\label{tab:population_largest}
\end{table}

On CIFAR-100, restricting \method{} to four cached points aligned separately
with the operating points effectively served by HeteroFL or ScaleFL retains
approximately 96\% and 98\% of the corresponding population margins. The
full 56-point cache adds only 1.08 and 0.53 accuracy points, respectively.
This control isolates serving density; the residual margins still reflect
complete-method differences in training and model space
(Supplementary Sec.~E).

\vspace{-4pt}

\paragraph{Post-hoc extraction.}
Fourteen uncached CIFAR-100 variants in the 25M--52M region have a mean
absolute gap of 1.14 points from the cached curve, and the 30M variant reaches
64.35\%, without further federated training. Reusing the architecture set on
CINIC-10 and TinyImageNet gives gaps of 1.34 and 1.79 points, respectively
(scope and extraction checks in Supplementary Sec.~E).

\subsection{Training Mechanisms and Controls}
\label{sec:training_rule_results}
\label{sec:sfn_dfn_results}
\label{sec:communication_results}
\label{sec:robustness_results}

\paragraph{Training rule.}
Endpoint-only rules perform poorly away from their trained endpoints
(Table~\ref{tab:component_ablation}). Relative to co-training both endpoints
with KD, full \method{} raises the 50M, 229M, 408M, and 596M variants by
5.17, 9.43, 8.22, and 0.97 points while remaining within 0.13 points at
25M, isolating the contribution of random-intermediate coverage.

\begin{table}[t]
\centering
\scriptsize
\setlength{\tabcolsep}{2.0pt}
\begin{tabular*}{\columnwidth}{@{\extracolsep{\fill}}lccccc@{}}
\toprule
Training rule & 25M & 50M & 229M & 408M & 596M \\
\midrule
Global min. only & 61.71 & 47.63 & 34.68 & 33.12 & 33.46 \\
Local max. only & 30.87 & 37.81 & 57.46 & 60.62 & 64.99 \\
Global min. + local max. & 64.14 & 60.25 & 61.90 & 62.55 & 70.36 \\
Global min. + local max. + KD & \textbf{65.02} & 60.43 & 62.39 & 63.72 & 70.09 \\
Full \method{} & 64.89 & \textbf{65.60} & \textbf{71.82} & \textbf{71.94} & \textbf{71.06} \\
\bottomrule
\end{tabular*}
\caption{CIFAR-100 component analysis under otherwise identical settings.
Global min. denotes the cache-wide minimum variant, whereas local max. denotes
each client's largest affordable cached variant. Bold marks the highest
accuracy in each column.}
\label{tab:component_ablation}
\end{table}

\begin{table}[t]
\centering
\scriptsize
\setlength{\tabcolsep}{1.3pt}
\begin{tabular*}{\columnwidth}{@{\extracolsep{\fill}}lccccc@{}}
\toprule
Training strategy & 25M & 50M & 229M & 408M & 596M \\
\midrule
SFN, unconstrained
& 1.05 & 1.01 & 0.92 & 2.32 & 5.19 \\
DFN, unconstrained
& 1.05 & 1.06 & 1.25 & 3.11 & 4.46 \\
SFN, budget-constrained
& 1.03 & 1.21 & 2.61 & 9.40 & 16.76 \\
DFN, budget-constrained
& 1.00 & 1.09 & 3.00 & 10.86 & 17.09 \\
\method{}
& \textbf{64.89} & \textbf{65.60} & \textbf{71.82}
& \textbf{71.94} & \textbf{71.06} \\
\bottomrule
\end{tabular*}
\caption{CIFAR-100 transfer results with each method's sampler and MaxNet
cosine-annealed, overlap-aware aggregation. Constrained variants additionally
use our client budgets and $\gamma=1$ allocation. Random classification
accuracy is 1\%.}
\label{tab:sfn_dfn_main}
\end{table}

Table~\ref{tab:sfn_dfn_main} tests whether the training procedures of the two
closest federated-supernet predecessors remain effective when transferred to
the 25M--596M shared model space and evaluated under our common federation
protocol. We retain each sampler together with MaxNet's cosine-annealed,
overlap-aware aggregation, so the comparison is a cross-setting stress test
rather than a native-setting reproduction. All four transfers remain near
chance at 25M; at 596M, the strongest reaches 17.09\% versus \method{}'s
71.06\%, a 53.97-point difference. The result shows that these predecessor
training procedures do not automatically extend to the expanded shared model
space and client-heterogeneity setting, motivating \method{}'s local
multi-variant co-training rule; performance in their native settings is
outside the comparison's scope.

\vspace{-4pt}
 
\paragraph{Training computation and parameter footprint.}
Under the analytic forward-plus-backward convolution-MAC proxy,
\method{} totals 31.01 PMACs, $1.56\times$ its single-variant reference and
$2.07\times$, $2.04\times$, and $1.03\times$ the HeteroFL, ScaleFL, and
FIARSE proxies. Matching this total gives baseline population averages of
44.44\%, 39.23\%, and 55.42\%, leaving \method{}'s 67.74\% ahead by
23.3, 28.5, and 12.3 points with 100/100 client wins. At 596M,
\method{}'s training envelope is 35.60M parameters versus
11.22M--11.27M baseline payloads; its deployed model is 11.78M.
Supplementary Sec.~E gives proxy definitions and full footprint tables.

\vspace{-4pt}

\paragraph{Communication and sensitivity.}
Relative to transmitting its full 55.68M-parameter supernet, \method{}
routing reduces the mean model-parameter payload to 8.17M parameters and
aggregate model-parameter traffic by $6.8\times$; per-client-round
model-parameter reductions average $43.8\times$ (median $22.9\times$,
maximum $111\times$). Across $s\in[0.8,1.5]$, 596M accuracy remains
70.29--74.54\%. Across four
Dirichlet settings, \method{} beats FIARSE at all 20 matched anchors, with
25M-to-596M changes of 5.77--7.37 versus 21.25--23.25 points
(full results in Supplementary Sec.~E).


\section*{Acknowledgment}
The computations in this work were enabled by resources provided by the
National Academic Infrastructure for Supercomputing in Sweden (NAISS),
funded by the Swedish Research Council.

\section{Conclusion}
\label{sec:conclusion}

Federated supernet training offers a train-once, deploy-many route for client
populations with heterogeneous deployment budgets, but our cross-setting stress
test shows that the evaluated predecessor training procedures do not
extend to a wider, budget-conditioned shared space. That broader
setting creates unequal
training exposure: parameters exclusive to high-cost variants are reachable
only by the high-budget minority, and their effective training signal also
depends on how data are distributed across budgets. We make this dependence
explicit through the $\gamma$-controlled allocation protocol and formulate the
setting through an elastic parameter space with client-specific affordability
and nested parameter reachability. \method{} addresses the resulting
optimization and communication challenges through local multi-variant
co-training, in-place distillation, budget-tailored sub-supernet routing, and
position-wise sample-count-weighted aggregation.

Across CIFAR-100, CINIC-10, and TinyImageNet-200, \method{} achieves the
strongest population-averaged accuracy among the evaluated weight-sharing
methods under largest-affordable serving. 
Sub-supernet routing substantially reduces aggregate model-parameter traffic
relative to full-supernet transmission. Future work should reduce the
overhead of local multi-variant co-training and test the formulation on larger
architectures and additional modalities.

\bibliography{ref}

\clearpage


\section*{Supplementary material}

# Supplementary Material for FEAST: Federated Shared-Space Training for Resource-Heterogeneous Clients

**Bostan Khan, Masoud Daneshtalab**
Mälardalen University, Västerås, Sweden

**Supplementary overview.** This supplementary material provides the complete protocol, implementation, experimental, and theoretical details supporting the main paper. Section A gives the client-budget and partition construction; Sec. B describes the shared model space, fixed variant cache, and BN-calibration protocol; Sec. C gives sub-supernet routing, sparse aggregation, and communication analysis; Sec. D provides the complete local co-training procedure and training-computation model; Sec. E reports additional experimental details, post-hoc extraction, cost controls, parameter footprints, and sensitivity results; and Sec. F provides the structural and finite-horizon analysis.

## A Client Budget Generation and Partition Realization

**Client population and MAC budgets.** Client $i$ receives a budget $b_i$, defined as the maximum per-example inference MAC cost of any individual variant that it may train or deploy. We draw its resource rank from

$$\Pr[\text{rank} = r] = \frac{r^{-s}}{\sum_{\ell=1}^{R} \ell^{-s}}, \qquad r \in \{1, \dots, R\}, \quad (1)$$

with $R = N$ and $s = 1.2$, following the moderate-skew setting used by Pisces (Jiang et al. 2022). Ranks represent budget levels rather than client identities, so several clients may draw the same rank.

We map rank $r_i$ to a jittered continuous budget in $[B_{\min}, B_{\max}] = [25\text{M}, 1500\text{M}]$:

$$\Delta_B = \frac{B_{\max} - B_{\min}}{R - 1}, \qquad \epsilon_i \sim \mathcal{U}\left(-\frac{\Delta_B}{2}, \frac{\Delta_B}{2}\right), \quad (2)$$
$$b_i = \operatorname{clip}(B_{\min} + (r_i - 1)\Delta_B + \epsilon_i; B_{\min}, B_{\max}),$$

where $\operatorname{clip}(x; L, U) = \min\{U, \max\{L, x\}\}$. The raw budget $b_i$ controls variant affordability. Data allocation uses $B_{\text{cap}} = 600\text{M}$ MACs, the approximate served-model ceiling, so capability above the served range does not further increase allocation weight.

**$\gamma$-allocation.** Define

$$q_i(\gamma) = \frac{\left(b_i^{\text{alloc}}\right)^{\gamma}}{\sum_{j=1}^{N} \left(b_j^{\text{alloc}}\right)^{\gamma}}, \qquad b_i^{\text{alloc}} = \min(b_i, B_{\text{cap}}), \quad (3)$$

where $\gamma \geq 0$. Here, $\gamma = 0$ gives uniform allocation weights and $\gamma = 1$ gives weights proportional to capped budgets.

**Budget-aware Dirichlet partition.** Let $N_{\text{train}}$ denote the client-training pool after excluding server-held data, and let class $k$ have a shuffled index set $\mathcal{I}_k$ of size $N_k$. We draw

$$\widetilde{\mathbf{p}}_k \sim \operatorname{Dir}(c_{k,1}, \dots, c_{k,N}), \qquad c_{k,i} = \alpha_{\text{data}} N_k q_i(\gamma). \quad (4)$$

Before assigning class $k$, the implementation sets $\widetilde{p}_{k,i} = 0$ for clients whose accumulated count has reached $q_i(\gamma)N_{\text{train}}$, and renormalizes the remaining entries to obtain $\mathbf{p}_k$. It then forms the first $N - 1$ integer split points

$$t_{k,i} = \left\lfloor N_k \sum_{j=1}^{i} p_{k,j} \right\rfloor, \qquad i = 1, \dots, N-1, \quad (5)$$

and partitions the shuffled indices contiguously. The final segment extends to the end of $\mathcal{I}_k$, so every example is assigned exactly once.

The quantity $q_i(\gamma)N_{\text{train}}$ is a pre-class balancing cap, not an exact realized target. Whole-class segments can move a client above the cap, while masking and renormalization can leave another client below it and introduce class-order dependence. The complete partition is redrawn if any client receives fewer than ten examples.

Before masking,

$$\mathbb{E}[\widetilde{p}_{k,i}] = \frac{c_{k,i}}{\sum_{j=1}^{N} c_{k,j}} = q_i(\gamma), \quad (6)$$

whereas the final integer shares can differ because of balancing, sequential class processing, integer splitting, and rejection resampling. The initial draw has total concentration

$$s_k = \sum_{i=1}^{N} c_{k,i} = \alpha_{\text{data}} N_k. \quad (7)$$

Thus, changing $\gamma$ changes the target mean vector while preserving total per-class precision, although component-level variability changes with $q_i(\gamma)$.

For comparison with a conventional symmetric Dirichlet over $N$ clients, we report

$$\alpha_{\text{eff},k} = \frac{s_k}{N} = \frac{\alpha_{\text{data}} N_k}{N},$$
$$\alpha_{\text{eff}} = \frac{\alpha_{\text{data}} N_{\text{train}}}{KN} \quad \text{for a balanced } K\text{-class dataset.} \quad (8)$$

The CIFAR-100 realized client-size and local-step ranges are reported in Sec. F.5.

# B Federated Shared Model Space and Fixed Variant Cache

**Elastic model and search space.** The exact architecture choices for the 55.68M-parameter OFA-style ResNet described in the main paper are as follows. The model contains $S = 4$ elastic stages with base channels $[128, 256, 512, 1024]$ and choices

$$\begin{aligned} d_\ell &\in \{0, \ldots, 8\}, & \ell &= 1, \ldots, S, \\ e_\ell &\in \{0.10, 0.14, 0.18, 0.22, 0.25\}, & \ell &= 1, \ldots, S, \\ w_j &\in \{0, \ldots, 9\}, & j &= 0, \ldots, S. \end{aligned} \tag{9}$$

Here, $d_\ell$ is the additional stage depth, $e_\ell$ is a stage-level expansion ratio, and $w_j$ indexes width multipliers $\{0.1, 0.2, \ldots, 1.0\}$ for the stem and four stages. These choices define the architecture space $\mathcal{H}$.

**Fixed cache and offline construction.** FEAST uses a fixed cache $\mathcal{A} \subset \mathcal{H}$ of 56 architectures spanning approximately 25M–600M MACs. For each target budget $B_\nu$, the Pareto-guided DeepFedNAS procedure (Khan and Daneshtalab 2026b,a) selects

$$\begin{aligned} \mathbf{a}^\star_\nu &= \arg\max_{\mathbf{a} \in \mathcal{H}} F_{\mathrm{DFN}}(\mathbf{a}) \\ \text{s.t.} \quad & \mathrm{MAC}(\mathbf{a}) \le B_\nu, \qquad \rho(\mathbf{a}) \le \rho_{0,\nu}. \end{aligned} \tag{10}$$

The cited work defines the training-free fitness $F_{\mathrm{DFN}}$ and effectiveness measure $\rho$. We reuse this offline construction procedure at the deployment budgets studied here; federation itself performs no architecture search.

Client $i$ can activate

$$\mathcal{A}_i = \{\mathbf{a} \in \mathcal{A} \mid \mathrm{MAC}(\mathbf{a}) \le b_i\}. \tag{11}$$

Affordability uses the raw budget $b_i$, whereas $b_i^{\mathrm{alloc}}$ affects only data allocation.

## B.1 Additional cache-generation details

Expansion is selected once per stage. Each target uses a genetic algorithm with population 256, 512 generations, mutation probability 0.3, feasibility-preserving initialization and mutation, tournament selection, single-point crossover, and elitism. Table 1 summarizes the main search and low-compute extension.

MACs are recomputed under each dataset's active input and stem configuration; the TinyImageNet cache therefore spans 27M–601M MACs, explaining its slightly higher largest operating point.

**BN calibration for evaluation.** For every evaluated operating point, BN running statistics are reset and recalibrated with the same server-held samples and image-level transformations in a fixed order; batch-level Mixup and CutMix are disabled during calibration.

| Segment | Realized range | Variants | $\rho_0$ | Pop./gen. |
|---|---|---|---|---|
| Main | 50M–596M | 50 | 0.31 | 256/512 |
| Low-compute extension | 25M–75M | 6 | 0.40 | 256/512 |
| Merged and MAC-sorted | 24.96M–595.67M | 56 | – | – |

Table 1: Construction of the fixed CIFAR-100/CINIC-10 variant cache. The effectiveness ceiling is relaxed for the low-compute extension to make targets below 50M MACs feasible; the search segment extends to 75M to provide overlap with the main cache during merging.

# C Sub-Supernet Construction, Sparse Aggregation, and Communication Analysis

**Routing envelope and construction.** The server samples a cohort $\mathcal{S}_t$ of $m$ clients in round $t$ and derives client-specific depth and width bounds from $\mathcal{A}_i$:

$$\begin{aligned} d^*_{i,\ell} &= \max_{\mathbf{a} \in \mathcal{A}_i} d_\ell(\mathbf{a}), \quad \ell = 1, \ldots, S, \\ w^*_{i,j} &= \max_{\mathbf{a} \in \mathcal{A}_i} w_j(\mathbf{a}), \quad j = 0, \ldots, S. \end{aligned} \tag{12}$$

The maxima may come from different cached variants, so the routed envelope can represent uncached combinations, although local sampling remains restricted to $\mathcal{A}_i$. No separate expansion bound is used: each retained bottleneck supports the globally allowed expansion choices at its routed width, and a selected $e_\ell$ activates a leading channel slice at runtime.

Global weights are copied by leading-dimension slicing. For $W \in \mathbb{R}^{C_{\mathrm{out}} \times C_{\mathrm{in}} \times k \times k}$, the client receives

$$W_i = W[\,:C^*_{\mathrm{out}},\ :C^*_{\mathrm{in}}, :, :], \tag{13}$$

where the retained channel counts are implied by the routed width bounds. Batch-normalization, bias, and linear-layer tensors are sliced analogously. The server also stores a map $\Omega_i$ from each local tensor position to its corresponding position in the global supernet. Suppressing the round index, we denote the routed state by $\boldsymbol{\theta}_i$; with routing disabled, $\boldsymbol{\theta}_i = \boldsymbol{\theta}^t$ and $\Omega_i$ is the identity map.

**Sparse sample-weighted aggregation.** Let $\mathcal{U}_i$ be the global positions covered by $\Omega_i$, and let $\bar{\boldsymbol{\theta}}'_i[u]$ be client $i$'s returned value after mapping to global position $u$. The contributors are

$$\mathcal{S}_t(u) = \{i \in \mathcal{S}_t \mid u \in \mathcal{U}_i\}. \tag{14}$$

The global update is

$$\boldsymbol{\theta}^{t+1}[u] = \begin{cases} \dfrac{\sum_{i \in \mathcal{S}_t(u)} |\mathcal{D}_i|\, \bar{\boldsymbol{\theta}}'_i[u]}{\sum_{i \in \mathcal{S}_t(u)} |\mathcal{D}_i|}, & \mathcal{S}_t(u) \neq \emptyset, \\ \boldsymbol{\theta}^t[u], & \mathcal{S}_t(u) = \emptyset. \end{cases} \tag{15}$$

The implementation realizes Eq. (15) with value and sample-count accumulators. A position is normalized only over returned sub-supernets that contain it; if none contains it, the previous global value is retained.

| Metric | Full FEAST | Routed FEAST |
|---|---|---|
| Mean parameters per client | 55.68M | 8.17M |
| Total traffic reduction | 1.0× | 6.8× |
| Mean client reduction | 1.0× | 43.8× |
| Median client reduction | 1.0× | 22.9× |
| Largest client reduction | 1.0× | 111× |

Table 2: FEAST-internal communication under the Zipf($s = 1.2$) client population. Reduction factors are measured relative to sending FEAST's full supernet to every selected client.

**Communication accounting.** Let $P$ be the full-supernet parameter count and $P_i$ the routed count for client $i$. Ignoring metadata and using the same precision, full-supernet round-trip communication transmits $2mP$ parameters, whereas routing transmits $2\sum_{i\in\mathcal{S}_t} P_i$. The reduction factor is

$$R_t^{\mathrm{comm}} = \frac{mP}{\sum_{i\in\mathcal{S}_t} P_i}. \tag{16}$$

For client $i$, the corresponding factor is $P/P_i$. These ratios use FEAST's full supernet as the reference payload.

Table 2 reports both the ratio of aggregate payloads and the distribution of client-level reduction ratios. The former is a ratio of sums, whereas the latter summarizes individual ratios.

# D Complete Local Multi-Variant Co-Training Procedure

**Variant selection.** Each selected client trains up to three cached variants on every local mini-batch: the global minimum, its local maximum, and a randomly sampled affordable intermediate when one exists. The endpoints are

$$\begin{aligned} \mathbf{a}^{\min} &= \arg\min_{\mathbf{a}\in\mathcal{A}} \mathrm{MAC}(\mathbf{a}), \\ \mathbf{a}_i^{\max} &= \arg\max_{\mathbf{a}\in\mathcal{A}_i} \mathrm{MAC}(\mathbf{a}). \end{aligned} \tag{17}$$

Because the client-budget range begins at the served floor, $\mathbf{a}^{\min} \in \mathcal{A}_i$ for every client.

For local mini-batch $\tau$, define

$$\mathcal{A}_i^{\mathrm{int}} = \{\mathbf{a}\in\mathcal{A}_i \mid \mathrm{MAC}(\mathbf{a}^{\min}) < \mathrm{MAC}(\mathbf{a}) < \mathrm{MAC}(\mathbf{a}_i^{\max})\}. \tag{18}$$

When nonempty, the client resamples

$$\mathbf{a}_{i,\tau}^{\mathrm{rand}} \sim \mathrm{Uniform}(\mathcal{A}_i^{\mathrm{int}}) \tag{19}$$

for each mini-batch. If $\mathbf{a}_i^{\max} = \mathbf{a}^{\min}$, the step reduces to supervised training of the global minimum.

**Local objective and distillation.** Each mini-batch produces one mixed-label tuple $(\tilde{x}, y_a, y_b, \lambda_{\mathrm{mix}})$, reused for every applicable variant. Let $z_{i,\tau}(\mathbf{a}) = f(\tilde{x};\boldsymbol{\theta}_i,\mathbf{a})$ and

$$\mathcal{L}_{\mathrm{sup}}(z) = \lambda_{\mathrm{mix}}\mathcal{L}_{\mathrm{CE}}(z, y_a) + (1-\lambda_{\mathrm{mix}})\mathcal{L}_{\mathrm{CE}}(z, y_b). \tag{20}$$

For distinct endpoints, the local maximum is evaluated first with $\mathcal{L}_{\max} = \mathcal{L}_{\mathrm{sup}}(z_{i,\tau}(\mathbf{a}_i^{\max}))$. Its detached logits define

$$\hat{p}_{i,\tau}^{\max} = \mathrm{softmax}(\mathrm{sg}(z_{i,\tau}(\mathbf{a}_i^{\max}))), \tag{21}$$

where sg denotes stop-gradient. The KD loss is

$$\mathcal{L}_{\mathrm{KD}}(z,\hat{p}) = -\sum_c \hat{p}_c \log \mathrm{softmax}(z)_c, \tag{22}$$

with implicit mini-batch averaging and no temperature scaling. Students use

$$\mathcal{L}_{\mathrm{stu}}(z,\hat{p}) = (1-\rho_{\mathrm{kd}})\mathcal{L}_{\mathrm{sup}}(z) + \rho_{\mathrm{kd}}\mathcal{L}_{\mathrm{KD}}(z,\hat{p}), \tag{23}$$

where $\rho_{\mathrm{kd}} = 0.5$. Thus, $\mathcal{L}_{\min} = \mathcal{L}_{\mathrm{stu}}(z_{i,\tau}(\mathbf{a}^{\min}), \hat{p}_{i,\tau}^{\max})$ and, when sampled, $\mathcal{L}_{\mathrm{rand}} = \mathcal{L}_{\mathrm{stu}}(z_{i,\tau}(\mathbf{a}_{i,\tau}^{\mathrm{rand}}), \hat{p}_{i,\tau}^{\max})$.

The mini-batch objective is

$$\mathcal{L}_{i,\tau} = \begin{cases} \mathcal{L}_{\mathrm{sup}}(z_{i,\tau}(\mathbf{a}^{\min})), & \mathbf{a}_i^{\max} = \mathbf{a}^{\min}, \\ \mathcal{L}_{\max} + \mathcal{L}_{\min}, & \mathcal{A}_i^{\mathrm{int}}=\emptyset,\ \mathbf{a}_i^{\max}\neq\mathbf{a}^{\min}, \\ \mathcal{L}_{\max} + \mathcal{L}_{\min} + \mathcal{L}_{\mathrm{rand}}, & \text{otherwise}. \end{cases} \tag{24}$$

The client backpropagates each applicable component, accumulates the resulting gradients, clips their combined $L_2$ norm to 10, and performs one optimizer step. The local maximum supplies detached soft targets to the smaller variants; the global minimum is trained by every selected client, and per-mini-batch intermediate sampling extends coverage across the affordable cache.

**Local training computation.** The budget $b_i$ constrains each activated variant separately, not the cumulative cost of sequential local passes. Let $\mathcal{T}_{i,t}$ be client $i$'s mini-batches in round $t$, let $n_{i,\tau}$ be the mini-batch size, and define

$$\mathcal{V}_{i,\tau} = \begin{cases} \{\mathbf{a}^{\min}\}, & \mathbf{a}_i^{\max} = \mathbf{a}^{\min}, \\ \{\mathbf{a}^{\min}, \mathbf{a}_i^{\max}\}, & \mathcal{A}_i^{\mathrm{int}}=\emptyset,\ \mathbf{a}_i^{\max}\neq\mathbf{a}^{\min}, \\ \{\mathbf{a}^{\min}, \mathbf{a}_{i,\tau}^{\mathrm{rand}}, \mathbf{a}_i^{\max}\}, & \mathcal{A}_i^{\mathrm{int}} \neq \emptyset. \end{cases} \tag{25}$$

The analytic forward-plus-backward convolution-MAC proxy is

$$\begin{aligned} C_{i,t}^{\mathrm{fwd}} &= \sum_{\tau\in\mathcal{T}_{i,t}} n_{i,\tau} \sum_{\mathbf{a}\in\mathcal{V}_{i,\tau}} \mathrm{MAC}(\mathbf{a}), \\ \widehat{C}_{i,t}^{\mathrm{train}} &= \kappa_{\mathrm{fb}} C_{i,t}^{\mathrm{fwd}}, \qquad \kappa_{\mathrm{fb}} = 3. \end{aligned} \tag{26}$$

The factor $\kappa_{\mathrm{fb}} = 3$ approximates backward convolution arithmetic as twice the forward pass. The proxy excludes optimizer, loss, augmentation, communication, aggregation, and other non-convolution operations, and is therefore not an end-to-end latency or energy measure. For a realized schedule, the federation proxy is $\sum_t \sum_{i\in\mathcal{S}_t} \widehat{C}_{i,t}^{\mathrm{train}}$. Expected totals average the random pass uniformly over $\mathcal{A}_i^{\mathrm{int}}$ and client participation under the common $m/N$ sampling model while retaining realized mini-batch sizes. Each active variant satisfies $\mathrm{MAC}(\mathbf{a}) \le b_i$, but cumulative local training is not bounded by $b_i$.

Algorithms 1 and 2 summarize the server and client operations, respectively.

# E Additional Experimental Details and Controls

This section preserves the complete supporting evidence referenced by the main Experiments section. It reports the

**Algorithm 1** FEAST: one federation round

**Require:** $\boldsymbol{\theta}^t$, $\mathcal{A}$, $\{(\mathcal{D}_i, b_i)\}_{i=1}^N$, $m$, $\rho_{\mathrm{kd}}$, and routing flag ROUTE
Sample client set $\mathcal{S}_t$ of size $m$
**for** each client $i \in \mathcal{S}_t$ **do**
  $\mathcal{A}_i \leftarrow \{\mathbf{a} \in \mathcal{A} \mid \mathrm{MAC}(\mathbf{a}) \leq b_i\}$
  Set $\mathbf{a}^{\min}$, $\mathbf{a}_i^{\max}$, and $\mathcal{A}_i^{\mathrm{int}}$ by Eqs. (17)–(18)
  **if** ROUTE **then**
    $(\boldsymbol{\theta}_i, \Omega_i) \leftarrow \textsc{CreateSubSupernet}(\boldsymbol{\theta}^t, \mathcal{A}_i)$ ▷ Sec. C
  **else**
    $\boldsymbol{\theta}_i \leftarrow \boldsymbol{\theta}^t$; $\Omega_i \leftarrow \mathrm{IdentityMap}(\boldsymbol{\theta}^t)$
  **end if**
  $\boldsymbol{\theta}_i' \leftarrow \textsc{ClientStep}(\boldsymbol{\theta}_i, \mathcal{D}_i, \mathbf{a}^{\min}, \mathbf{a}_i^{\max}, \mathcal{A}_i^{\mathrm{int}}, \rho_{\mathrm{kd}})$
**end for**
$\boldsymbol{\theta}^{t+1} \leftarrow \textsc{SparseAggregate}(\boldsymbol{\theta}^t, \{(\boldsymbol{\theta}_i', \Omega_i, |\mathcal{D}_i|)\}_{i \in \mathcal{S}_t})$ ▷ Sec. C
**return** $\boldsymbol{\theta}^{t+1}$

**Algorithm 2** FEAST: local multi-variant co-training

**Require:** $\boldsymbol{\theta}_i$, $\mathcal{D}_i$, $\mathbf{a}^{\min}$, $\mathbf{a}_i^{\max}$, $\mathcal{A}_i^{\mathrm{int}}$, and $\rho_{\mathrm{kd}}$
$\boldsymbol{\theta}_i \leftarrow \textsc{Copy}(\boldsymbol{\theta}_i)$
**for** each mini-batch $\tau$ with $(x, y) \in \mathcal{D}_i$ **do**
  Construct $(\tilde{x}, y_a, y_b, \lambda_{\mathrm{mix}})$
  Zero gradients
  **if** $\mathbf{a}_i^{\max} = \mathbf{a}^{\min}$ **then**
    Activate $\mathbf{a}^{\min}$; compute $\mathcal{L}_{\mathrm{sup}}$ by Eq. (20); backward
  **else**
    Activate $\mathbf{a}_i^{\max}$; compute $\mathcal{L}_{\max}$; backward
    Compute $\hat{p}_{i,\tau}^{\max}$ by Eq. (21)
    Activate $\mathbf{a}^{\min}$; compute $\mathcal{L}_{\min}$ by Eq. (23); backward
    **if** $\mathcal{A}_i^{\mathrm{int}} \neq \emptyset$ **then**
      Sample $\mathbf{a}_{i,\tau}^{\mathrm{rand}}$ by Eq. (19)
      Activate $\mathbf{a}_{i,\tau}^{\mathrm{rand}}$; compute $\mathcal{L}_{\mathrm{rand}}$ by Eq. (23); backward
    **end if**
  **end if**
  Clip gradients to $L_2$ norm 10; optimizer step
**end for**
**return** $\boldsymbol{\theta}_i$

dataset and operating-point details, the exact SFN/DFN transfer configuration, the scope of post-hoc extraction, training-computation and parameter-footprint controls, and the full CIFAR-100 sensitivity sweeps.

## E.1 Datasets, Calibration, and Operating Points

Table 3 gives the dataset-specific partition and calibration sizes. Training counts exclude the server-held validation set. For each evaluated operating point, BN statistics are recalibrated using the disjoint server-held calibration samples described in Sec. B. All methods train for 4000 communication rounds, with 10 of 100 clients sampled uniformly without replacement per round. Each selected client performs one local epoch with batch size 64. SGD uses initial learning rate 0.025, momentum 0.9, cosine decay over the 4000-round horizon, and gradient clipping to $L_2$ norm 10. Weight decay is $10^{-4}$ except for FIARSE, which retains its original zero-weight-decay setting. Training uses $\mathrm{RandAugment}(2, 6)$ and alternates Mixup with parameter 0.4 and CutMix with parameter 1.0. Validation is evaluated every 20 rounds to select the checkpoint; the test set is not used for checkpoint selection.

For CIFAR-100, 45,000 training examples are partitioned among clients and 5,000 are server-held, comprising 500 BN-calibration and 4,500 validation examples. For TinyImageNet, the corresponding counts are 90,000 client, 1,000 calibration, and 9,000 validation examples. CINIC-10 partitions all 90,000 examples from its training directory among clients and uses its separate 90,000-example validation directory for 9,000 calibration and 81,000 checkpoint-selection examples. The official test splits remain disjoint. After selecting the best-validation checkpoint, every evaluated operating point receives its own deterministic BN recalibration before the single final test evaluation.

Table 4 summarizes the deployment structures used for the main comparison. FEAST is evaluated on all 56 cached variants, HeteroFL and ScaleFL on their fixed native levels, and FIARSE on measured points from its continuous mask family. No accuracy values are interpolated. For largest-affordable population evaluation, FIARSE is evaluated on a dense MAC grid because its mask family supports intermediate operating points.

| Dataset | Classes | Resolution | Train/Test | $\alpha_{\mathrm{data}}$ | BN cal. |
|---|---|---|---|---|---|
| CIFAR-100 | 100 | 32×32 | 45K/10K | 0.1 | 500 |
| CINIC-10 | 10 | 32×32 | 90K/90K | 0.005 | 9000 |
| TinyImageNet | 200 | 64×64 | 90K/10K | 0.1 | 1000 |

Table 3: Dataset-specific settings. BN cal. is the number of disjoint server-held samples used to recalibrate each evaluated operating point.

**Baseline adaptations and assignment.** HeteroFL, ScaleFL, and FIARSE use a common CIFAR-style ResNet substrate with hidden widths $[64, 128, 256, 512]$, four stages, and two BasicBlocks per stage. The full-width models cost approximately 550M–555M MACs. CIFAR-100 and CINIC-10 use a stride-one stem without max pooling, while TinyImageNet uses its dataset-specific stem configuration. These adaptations align the methods' compute ranges while retaining their defining mechanisms.

HeteroFL (Diao, Ding, and Tarokh 2021) retains width multipliers $\{1/16, 1/8, 1/4, 1/2, 1\}$, top-left channel slicing, and position-wise aggregation. ScaleFL (Ilhan, Su, and Liu 2023) retains four depth–width levels with $(\mathrm{blocks}, \mathrm{width})$ equal to $(3, 0.40)$, $(4, 0.50)$, $(5, 0.60)$, and $(8, 1.00)$, together with its early-exit self-distillation. FIARSE (Wu et al. 2024) retains its magnitude-based unstructured masks, threshold-controlled biased gradient descent, and sparse-delta aggregation. All methods use the same client partitions, resource budgets, participation schedule, and augmentation protocol.

| Method | Deployment structure | MAC range |
|---|---|---|
| FEAST | 56 cached variants | 25M–601M |
| HeteroFL | 5 width tiers | 2M–550M |
| ScaleFL | 4 depth–width levels | 35M–555M |
| FIARSE | continuous masks | 9M–555M |

Table 4: Native deployment structures of the four weight-sharing methods. The upper FEAST value is 601M on TinyImageNet and 596M on CIFAR-100 and CINIC-10.

| Quantity | HeteroFL-aligned | ScaleFL-aligned |
|---|---|---|
| FEAST subset (M MACs) | 25.0/34.6/139.0/549.4 | 34.6/71.7/128.2/549.4 |
| FEAST, full 56 points | 67.74 | 67.74 |
| FEAST, matched subset | 66.66 | 67.21 |
| Baseline, native tiers | 41.50 | 35.78 |
| Full margin | 26.24 | 31.96 |
| Density contribution | 1.08 | 0.53 |
| Residual margin | 25.16 | 31.43 |
| FEAST client wins | 100/100 | 100/100 |

Table 5: CIFAR-100 serving-density control under largest-affordable population evaluation. Accuracies and margins are percentages and percentage points, respectively. Differences are computed from the displayed values; density contribution is full-curve minus matched-subset accuracy.

For population evaluation, each client receives the largest measured native operating point whose MAC cost does not exceed its budget. If no native point is affordable, the method's smallest trained point is used as a minimum-tier fallback. This occurs only for ScaleFL: 32 clients on CIFAR-100 and CINIC-10, and 31 clients on TinyImageNet, have budgets below its approximately 35M-MAC minimum and therefore receive level 1. The ScaleFL population result should accordingly be read as largest-affordable serving with a minimum-tier fallback for otherwise unserved clients.

**Operating-point density control.** To isolate the effect of denser serving, we repeat the CIFAR-100 population evaluation after retaining only the cached FEAST variant nearest to each native HeteroFL or ScaleFL tier in MACs. Because client budgets begin at 25M MACs, HeteroFL's 2M and 9M tiers both map to FEAST's 25M floor, yielding four unique FEAST points; ScaleFL also yields four. We use the same fixed client budgets and largest-affordable rule as in the main-paper population evaluation, without retraining or interpolating accuracy.

Table 5 shows that only 1.08 and 0.53 points arise from density, leaving approximately 96% and 98% of the full margins. These residuals compare complete methods, not training rules alone, because model-space and cache-design differences remain. FIARSE is already evaluated densely.

## E.2 SFN and DFN Transfer Configuration

The transferred SuperFedNAS (SFN) procedure uses its uniform random architecture sampler, whereas DeepFedNAS (DFN) uses its Pareto-path sampler. Both retain MaxNet's overlap-aware aggregation, with its coefficient annealed by a cosine schedule from 0.9 to 0.125. The sampler and aggregation are therefore transferred together, rather than replacing MaxNet aggregation with ordinary FedAvg.

All four reported transfers use the canonical augmentation, optimizer, 100-client, 10%-participation, and 4000-round settings from the main paper. The unconstrained variants sample from the complete expanded supernet. The budget-constrained variants additionally use the Zipf($s = 1.2$) client budgets and $\gamma = 1$ allocation, and restrict each client to affordable architectures. The comparison asks whether each method's sampler and MaxNet aggregation remain effective when transferred to the 25M–596M shared model space and evaluated under the federation protocol described in the main paper. It is a cross-setting stress test, not a reproduction in the methods' original model spaces or client settings, and conclusions about performance in those native settings are outside its scope.

## E.3 Post-Hoc Extraction

After federation, FEAST can search for an architecture under a requested MAC limit and extract its weights from the trained supernet. We use the training-free objective in Eq. (10) with population 512, 512 generations, mutation probability 0.3, and the cache-construction thresholds $\rho_0 = 0.40$ below 50M MACs and $\rho_0 = 0.31$ otherwise. Neither validation nor test accuracy enters selection. The extracted subnet receives architecture-specific BN recalibration, but no weight fine-tuning or further federated training.

We searched 29 target budgets spanning approximately 25M–560M. The low-end results from 25M-52M MAC underlying the main-paper values use 14 logged GA outputs. Architectures discovered on CIFAR-100 are reused on CINIC-10 and TinyImageNet, with MACs recomputed for TinyImageNet. Gaps are measured against the linearly interpolated cached-variant curve at the same realized MAC value.

| Dataset | Evaluation range | Mean $\lvert\text{gap}\rvert$ (pp) |
|---|---|---|
| CIFAR-100 | Low end: 25.54M–51.32M | 1.14 |
| CIFAR-100 | Full: 25.54M–557.91M | 0.68 |
| CINIC-10 | Low end: 25.53M–51.31M | 1.34 |
| CINIC-10 | Full: 25.53M–557.83M | 0.88 |
| TinyImageNet | Low end: 26.87M–54.65M | 1.79 |
| TinyImageNet | Full: 26.87M–563.30M | 1.03 |

Table 6: Mean absolute post-hoc extraction gap from the linearly interpolated cached-variant curve at the same MAC value. Low-end results use 14 GA outputs; full-range results use 29.

At 30M MACs, the search subnet reaches 64.35% on CIFAR-100, only 0.33 points below the cached-curve reference. Together with Table 6, this shows that the trained supernet can serve additional architectures across the deployment range without further federated training.

**Client-side subnet extraction.** We also compared extraction from routed 50M and 229M client sub-supernets against extraction from the complete server supernet. For cached variants, the minimum, an intermediate, and the largest affordable architecture at each budget matched across all 2352

| Method | Logical PMAC | Executed PMAC | FEAST/executed |
|---|---|---|---|
| HeteroFL | 14.97 | 14.97 | $2.07\times$ |
| ScaleFL | 15.22 | 15.22 | $2.04\times$ |
| FIARSE | 19.12 | 29.97 | $1.03\times$ |
| Single-variant reference | 19.86 | 19.86 | $1.56\times$ |
| FEAST | 31.01 | 31.01 | $1.00\times$ |

Table 7: Expected convolution-MAC training proxies over 4000 rounds on CIFAR-100. The single-variant reference trains only each client's deployed variant under the FEAST population and partition.

tensor comparisons and all six accuracy measurements. Of the 14 logged low-end GA outputs, all fit the 229M payload and five fit the tighter 50M payload; every fitting output again matched full-supernet extraction exactly. Thus, client-side extraction is exact when the architecture lies within the routed payload's position-specific depth and width limits, although MAC affordability alone does not guarantee this structural fit.

## E.4 Training-Computation Controls

Equation (26) defines the analytic forward-plus-backward convolution-MAC proxy. It retains realized client shard sizes and the exact one-, two-, or three-variant local case, while averaging uniformly over the random intermediate and the common client-participation model. It excludes optimizer, loss, augmentation, communication, aggregation, and other non-convolution operations, and is not a latency or energy measurement.

Table 7 reports the resulting 4000-round totals. FIARSE is shown under both logical active arithmetic and executed dense-convolution arithmetic. Its unstructured mask acts on the full dense weight tensors, so the latter represents the computation executed by the implementation and is used for the matched-compute control.

Of the 100 FEAST clients, 31 use one variant per mini-batch, six use two, and 63 use three. The federation-total FEAST-to-single-variant ratio is $1.56\times$; the client-level ratios range from $1.00\times$ to $2.34\times$. Thus, the cumulative cost is lower than a uniform three-pass approximation because low-budget clients frequently use degenerate one- or two-variant cases.

**Equal-total-training-compute control.** We extend each baseline's training horizon so that its total executed convolution-MAC proxy matches FEAST's 31.01 PMACs, while retaining the same optimization schedule relative to that horizon. Table 8 reports the largest-affordable population results. For reference, under the original 4000-round training horizon, the largest-affordable population evaluation gives 67.74% for FEAST and 41.50%, 35.78%, and 55.13% for HeteroFL, ScaleFL, and FIARSE, respectively.

The native low- and high-compute operating points change from 48.33%/68.40% to 52.20%/70.23% for HeteroFL, from 20.98%/65.01% to 23.25%/67.29% for ScaleFL, and from 48.20%/68.65% to 48.22%/68.58% for FIARSE. Longer training therefore improves HeteroFL and ScaleFL modestly and has little effect on FIARSE, while the 12.33–28.51-point population margins remain. This control indicates that the primary margins are not explained solely by FEAST's greater convolution arithmetic.

| Method | Orig. r. | Match r. | Orig. | Match | Margin | Wins |
|---|---|---|---|---|---|---|
| HeteroFL | 4000 | 8286 | 41.50 | 44.44 | 23.30 | 100/100 |
| ScaleFL | 4000 | 8150 | 35.78 | 39.23 | 28.51 | 100/100 |
| FIARSE | 4000 | 4139 | 55.13 | 55.42 | 12.33 | 100/100 |

Table 8: Equal-total-training-compute control on CIFAR-100. Original and matched columns are population-averaged accuracy (%) under largest-affordable serving. Margin is relative to FEAST's 67.74%; wins counts client-level FEAST comparisons after compute matching. Displayed accuracies are rounded; margins use full-precision outputs.

| Method | 25M | 50M | 229M | 408M | 596M |
|---|---|---|---|---|---|
| FEAST routed envelope | 0.50M | 2.41M | 10.02M | 20.61M | 35.60M |
| HeteroFL | 0.18M | 0.71M | 2.82M | 2.82M | 11.22M |
| ScaleFL | 0.07M | 0.07M | 0.61M | 0.61M | 11.27M |
| FIARSE transmitted tensor | 11.22M | 11.22M | 11.22M | 11.22M | 11.22M |

Table 9: Client-side parameter counts during training at the five canonical CIFAR-100 budget anchors.

## E.5 Parameter Footprints

The server stores a 55.68M-parameter FEAST supernet, compared with 11.22M, 11.27M, and 11.22M parameters for the complete HeteroFL, ScaleFL, and FIARSE models, respectively. Client-side training and deployment require different quantities. Table 9 counts the transmitted training model. For FEAST, this is the routed envelope over all affordable cached variants; for HeteroFL and ScaleFL, it is the selected physical tier; and for FIARSE, it remains the complete dense tensor because masking does not reduce the transmitted parameter array.

At 596M, the routed FEAST envelope is 3.0–$3.2\times$ larger than the three baseline payloads. This envelope supports the global minimum, any sampled intermediate, and the local maximum within one routed model; it is not the single architecture deployed for inference.

Table 10 instead counts the single deployed model. The FEAST values are the parameters of the selected cached architecture. For FIARSE, they count active post-mask parameters rather than the transmitted dense tensor.

At 25M–408M, the deployed FEAST variant lies between the smaller HeteroFL/ScaleFL models and the larger FIARSE active model. At 596M, its 11.78M parameters are within approximately 5% of all three baselines. The large high-budget footprint is therefore specific to multi-variant training, not single-variant deployment.

## E.6 Resource and Label-Skew Sensitivity

Table 11 varies the Zipf exponent while holding $\gamma = 1$ and $\alpha_{\mathrm{data}} = 0.1$. Increasing $s$ places more clients at low budgets and predictably reduces high-budget accuracy, but no tested setting inverts the curve. The 596M result remains 70.29%–74.54%, and every higher-budget anchor exceeds the 25M

| Method | 25M | 50M | 229M | 408M | 596M |
|---|---|---|---|---|---|
| FEAST deployed variant | 0.33M | 0.72M | 4.02M | 7.99M | 11.78M |
| HeteroFL | 0.18M | 0.71M | 2.82M | 2.82M | 11.22M |
| ScaleFL | 0.07M | 0.07M | 0.61M | 0.61M | 11.27M |
| FIARSE active parameters | 0.51M | 1.02M | 4.64M | 8.25M | 11.22M |

Table 10: Parameters in the single model deployed at each budget.

| Zipf $s$ | 25M | 50M | 229M | 408M | 596M |
|---|---|---|---|---|---|
| 0.8 | 65.50 | 66.25 | 74.65 | 74.15 | 74.54 |
| 1.0 | 64.31 | 65.77 | 72.55 | 72.81 | 72.16 |
| 1.2 | 64.89 | 65.60 | 71.82 | 71.94 | 71.06 |
| 1.5 | 64.37 | 65.66 | 70.77 | 70.79 | 70.29 |

Table 11: CIFAR-100 sensitivity of FEAST to the resource-skew exponent.

result.

Table 12 compares FEAST and FIARSE under four nominal Dirichlet settings, with $s = 1.2$ and $\gamma = 1$ held fixed. $\alpha_{\mathrm{data}} \in \{0.05, 0.10, 0.30, 1.00\}$ corresponds to $\alpha_{\mathrm{eff}} \in \{0.225, 0.45, 1.35, 4.5\}$, respectively. FEAST exceeds FIARSE at all 20 matched anchor comparisons. Its 25M-to-596M change remains 5.77–7.37 points, versus 21.25–23.25 points for FIARSE, while its 596M margin ranges from 2.41 to 7.69 points. The result supports robustness over the tested label-skew range without implying invariance to the partition parameter.

**$\gamma$ sweep.** For completeness, Table 13 reports the full five-anchor FEAST sweep used by the main paper. The 596M result increases at every sampled value from $\gamma = 0$ through $\gamma = 1.5$, while the 25M-to-596M change progresses from $-4.29$ to $+8.40$ points. We use $\gamma = 1$ as the proportional cross-method default rather than claiming that it is optimal.

# F Structural and Finite-Horizon Analysis

This appendix isolates the theoretical claims needed to interpret the protocol and method contributions analyzed here. We first formalize the $\gamma$-allocation protocol independently of FEAST. We then characterize budget-dependent reachability, activation exposure, and sparse aggregation in the federated shared model space. Finally, an auxiliary frozen-gradient operator yields a routed computation scale that motivates a conditional finite-horizon bound for the full algorithm. Sparse-aggregation and implementation-specific effects are represented by bounded discrepancy terms; distillation, clipping, local drift, momentum, and weight decay are grouped rather than treated as separate theoretical contributions.

Main-paper Proposition 1 is established by Proposition F.2; Proposition 2 combines Proposition F.3, Corollary F.1, and Proposition F.4; and Theorem 1 is stated with exact constants and proved in Sec. F.4.

## F.1 Protocol allocation identity

The allocation protocol is a property of the simulated federation, not a component of FEAST. Let $\mathbf{b} = (b_1, \ldots, b_N)$

| $\alpha_{\mathrm{data}}$ | Method | 25M | 50M | 229M | 408M | 596M | Gap |
|---|---|---|---|---|---|---|---|
| 0.05 | FEAST | 64.19 | 64.89 | 72.06 | 72.60 | 71.56 | 7.37 |
| | FIARSE | 42.47 | 51.21 | 54.88 | 62.53 | 63.87 | 21.40 |
| 0.10 | FEAST | 64.89 | 65.60 | 71.82 | 71.94 | 71.06 | 6.17 |
| | FIARSE | 45.40 | 55.05 | 59.75 | 66.72 | 68.65 | 23.25 |
| 0.30 | FEAST | 65.09 | 65.28 | 72.03 | 72.24 | 71.42 | 6.33 |
| | FIARSE | 42.09 | 51.41 | 54.85 | 62.99 | 64.78 | 22.69 |
| 1.00 | FEAST | 64.66 | 65.19 | 70.77 | 70.74 | 70.43 | 5.77 |
| | FIARSE | 42.42 | 50.54 | 55.14 | 61.72 | 63.67 | 21.25 |

Table 12: CIFAR-100 accuracy (%) and 25M-to-596M change across nominal Dirichlet settings.

| $\gamma$ | 25M | 50M | 229M | 408M | 596M |
|---|---|---|---|---|---|
| 0 | 63.25 | 64.02 | 63.71 | 61.35 | 58.96 |
| 0.25 | 63.81 | 64.92 | 67.05 | 65.40 | 63.79 |
| 0.5 | 63.06 | 64.62 | 67.96 | 67.42 | 65.76 |
| 1.0 | 64.89 | 65.60 | 71.82 | 71.94 | 71.06 |
| 1.5 | 66.43 | 66.48 | 74.25 | 74.46 | **74.83** |

Table 13: FEAST sensitivity to $\gamma$ on CIFAR-100 under otherwise identical budgets, partition-generation settings, training, and evaluation.

denote the fixed client-budget vector, and let $\bar{b}_i = \min(b_i, B_{\mathrm{cap}})$ denote the budget used for data allocation.

**Definition F.1 ($\gamma$-allocation target)** *For $\gamma \geq 0$, define*

$$q_i(\gamma) = \frac{\bar{b}_i^{\gamma}}{\sum_{j=1}^{N} \bar{b}_j^{\gamma}}, \qquad \sum_{i=1}^{N} q_i(\gamma) = 1. \tag{27}$$

*For class $k$ with $N_k$ examples, the pre-integer allocation is*

$$\widetilde{\mathbf{p}}_k \sim \mathrm{Dir}\left(\alpha_{\mathrm{data}} N_k q_1(\gamma), \ldots, \alpha_{\mathrm{data}} N_k q_N(\gamma)\right). \tag{28}$$

**Proposition F.1 (Expected protocol allocation)** *For every class $k$ and client $i$,*

$$\mathbb{E}[\widetilde{p}_{k,i} \mid \mathbf{b}] = q_i(\gamma), \qquad \sum_{i=1}^{N} \alpha_{\mathrm{data}} N_k q_i(\gamma) = \alpha_{\mathrm{data}} N_k. \tag{29}$$

*Consequently, define the idealized pre-integer client count by*

$$\widetilde{n}_i = \sum_k N_k \widetilde{p}_{k,i}. \tag{30}$$

*Then*

$$\mathbb{E}[\widetilde{n}_i \mid \mathbf{b}] = N_{\mathrm{train}} q_i(\gamma). \tag{31}$$

**Proof.** Conditioned on the fixed client-budget vector $\mathbf{b}$, the concentrations $c_{k,i} = \alpha_{\mathrm{data}} N_k q_i(\gamma)$ are fixed. Using the mean of a Dirichlet component,

$$\mathbb{E}[\widetilde{p}_{k,i} \mid \mathbf{b}] = \frac{c_{k,i}}{\sum_{j=1}^{N} c_{k,j}} = \frac{\alpha_{\mathrm{data}} N_k q_i(\gamma)}{\alpha_{\mathrm{data}} N_k \sum_{j=1}^{N} q_j(\gamma)} = q_i(\gamma), \tag{32}$$

where the final equality uses $\sum_j q_j(\gamma) = 1$. The total concentration is

$$\sum_{i=1}^{N} c_{k,i} = \alpha_{\mathrm{data}} N_k \sum_{i=1}^{N} q_i(\gamma) = \alpha_{\mathrm{data}} N_k, \tag{33}$$

which proves Eq. (29). By the definition of $\widetilde{n}_i$, summing the expected class-wise counts over $k$ and using $\sum_k N_k = N_{\text{train}}$ gives Eq. (31). $\square$

The per-class total precision is $\alpha_{\text{data}} N_k$. Its symmetric-equivalent per-client concentration is therefore $\alpha_{\text{eff},k} = \alpha_{\text{data}} N_k / N$; for a balanced $K$-class dataset this reduces to the main-paper value $\alpha_{\text{data}} N_{\text{train}}/(KN)$. The case $\gamma = 0$ denotes the uniform-$q_i$ endpoint of this same budget-aware construction, not a separate strict symmetric $\text{Dir}(\alpha_{\text{data}})$ draw. Individual component concentrations and marginal variances depend on $q_i(\gamma)$. Thus, the protocol changes the target mean vector under fixed total per-class precision; it does not hold all higher-order allocation statistics fixed.

The implemented integer assignment and balancing constraints produce the realized data size $n_i$, which can differ from both $\widetilde{n}_i$ and its expectation in Eq. (31). All FEAST results below are therefore stated for the realized data sizes $\{n_i\}_{i=1}^N$. The idealized mean is used only when explicitly instantiating the protocol model.

## F.2 Reachability and activation exposure

Throughout the realized-partition analysis, assume that the client shards are disjoint and exhaustive, $\sum_{i=1}^N n_i = N_{\text{train}}$, that $n_i > 0$ for every client, and that every budget can afford the global-minimum cached variant. These conditions hold for the canonical partitions analyzed below.

Let $\mathcal{A}$ be the fixed variant cache and let

$$\mathcal{A}_i = \{\mathbf{a} \in \mathcal{A} : \text{MAC}(\mathbf{a}) \le b_i\} \tag{34}$$

be client $i$'s affordable set. The routed sub-supernet is the smallest implemented depth-width envelope containing all variants in $\mathcal{A}_i$, with each retained block supporting all allowed expansion choices; its global parameter positions are denoted by $\mathcal{U}_i$.

**Definition F.2 (Routed reachability)** *For a served position $u \in \mathcal{U}^{\text{srv}} = \bigcup_{i=1}^N \mathcal{U}_i$, define*

$$r_i(u) = \mathbb{I}\{u \in \mathcal{U}_i\}. \tag{35}$$

**Definition F.3 (Local activation exposure)** *On mini-batch $\tau$, let $\mathcal{V}_{i,\tau}$ be the set containing the global-minimum and local-maximum variants, with one uniformly sampled strict intermediate when available. Assume that this variant-set sampling rule is stationary across local mini-batches. Define*

$$a_{i,\tau}(u) = \mathbb{I}\left\{ u \in \bigcup_{\mathbf{a} \in \mathcal{V}_{i,\tau}} \boldsymbol{\theta}_{\mathbf{a}} \right\}, \tag{36}$$

$$h_i(u) = \mathbb{E}\left[ \sum_{\mathbf{a} \in \mathcal{V}_{i,\tau}} \mathbb{I}\{u \in \boldsymbol{\theta}_{\mathbf{a}}\} \,\middle|\, b_i \right]. \tag{37}$$

Here $r_i(u)$ determines aggregation eligibility, whereas $h_i(u)$ is the expected activation multiplicity per local optimization step and may exceed one. Coordinate activity satisfies $a_{i,\tau}(u) \le r_i(u)$. Sparse aggregation follows routed support, while supervised task gradients require coordinate activation. Tensor-level optimizer effects on routed but coordinate-inactive positions are absorbed into the algorithmic-discrepancy term used later.

**Proposition F.2 (Nested budget reachability)** *If $b_i \le b_j$, then*

$$\mathcal{A}_i \subseteq \mathcal{A}_j, \qquad \mathcal{U}_i \subseteq \mathcal{U}_j, \qquad r_i(u) \le r_j(u) \quad \textit{for every } u. \tag{38}$$

*Consequently, each served position is routed by an upper tail of the budget-ordered client population.*

**Proof.** Affordability is defined by a budget threshold, so increasing the budget cannot remove a cached variant. Each routed depth or width bound is a coordinate-wise maximum over the affordable set. Taking a maximum over a superset cannot reduce any bound. The retained blocks and channel dimensions are nested leading slices, and each retained block supports all allowed expansion choices. Therefore, monotone depth-width bounds imply $\mathcal{U}_i \subseteq \mathcal{U}_j$, which proves the indicator inequality. $\square$

For a cohort $\mathcal{S}_t$ of $m$ clients sampled uniformly without replacement, define the realized routed sample mass and the sample-weighted activation-exposure proxy by

$$M_t^{\text{route}}(u) = \sum_{i \in \mathcal{S}_t} n_i r_i(u), \tag{39}$$

$$M_t^{\text{exp}}(u) = \sum_{i \in \mathcal{S}_t} n_i h_i(u). \tag{40}$$

With a fixed mini-batch size and one local epoch, this proxy is proportional, up to batching and rounding, to the expected round-level activation count.

**Proposition F.3 (Expected exposure)** *Conditioned on the realized budgets and local data sizes,*

$$\mathbb{E}\left[M_t^{\text{route}}(u) \mid \{n_i, b_i\}_{i=1}^N\right] = \frac{m}{N} \sum_{i=1}^N n_i r_i(u), \tag{41}$$

$$\mathbb{E}\left[M_t^{\text{exp}}(u) \mid \{n_i, b_i\}_{i=1}^N\right] = \frac{m}{N} \sum_{i=1}^N n_i h_i(u). \tag{42}$$

*If $R_u = \sum_i r_i(u)$ clients route $u$, then*

$$\Pr[M_t^{\text{route}}(u) = 0] = \frac{\binom{N-R_u}{m}}{\binom{N}{m}}, \tag{43}$$

*with value zero when $N - R_u < m$.*

**Proof.** Each client has inclusion probability $m/N$. Linearity of expectation gives Eqs. (41) and (42). Equation (43) counts cohorts drawn entirely from the $N - R_u$ clients that do not route the position. $\square$

**Corollary F.1 (Protocol instantiation)** *Assume that cohort sampling is independent of the idealized allocation draw. Under the idealized pre-integer counts $\{\widetilde{n}_i\}_{i=1}^N$, define $\widetilde{M}_t^{\text{route}}(u)$ and $\widetilde{M}_t^{\text{exp}}(u)$ by replacing $n_i$ with $\widetilde{n}_i$ in Eqs.* (39) *and* (40)*. Then*

$$\mathbb{E}[\widetilde{M}_t^{\text{route}}(u) \mid \mathbf{b}] = \frac{mN_{\text{train}}}{N} \sum_{i=1}^N q_i(\gamma) r_i(u), \tag{44}$$

$$\mathbb{E}[\widetilde{M}_t^{\text{exp}}(u) \mid \mathbf{b}] = \frac{mN_{\text{train}}}{N} \sum_{i=1}^N q_i(\gamma) h_i(u). \tag{45}$$

**Proof.** Take expectation over the idealized allocation and substitute $\mathbb{E}[\widetilde{n}_i \mid \mathbf{b}] = N_{\text{train}} q_i(\gamma)$ into the definitions above. □

The corollary preserves the contribution boundary: $q_i(\gamma)$ belongs to the evaluation protocol, $h_i(u)$ belongs to FEAST's local co-training rule, and $r_i(u)$ belongs to FEAST's routing and aggregation geometry.

**Proposition F.4 ($\gamma$ increases routed allocation share)** *For a served position $u$, define*

$$W_u^{\text{route}}(\gamma) = \sum_{i=1}^{N} q_i(\gamma) r_i(u). \tag{46}$$

*Then $W_u^{\text{route}}(\gamma)$ is nondecreasing in $\gamma$.*

**Proof.** Let $x_i = \log \bar{b}_i$. Since $q_i(\gamma) = \exp(\gamma x_i)/\sum_j \exp(\gamma x_j)$,

$$\frac{\mathrm{d}}{\mathrm{d}\gamma} W_u^{\text{route}}(\gamma) = \operatorname{Cov}_{q(\gamma)}(r_i(u), x_i). \tag{47}$$

Its pairwise form is

$$\frac{1}{2}\sum_{i=1}^{N}\sum_{j=1}^{N} q_i(\gamma) q_j(\gamma) \big(r_i(u) - r_j(u)\big)\big(x_i - x_j\big). \tag{48}$$

The capped budget $\bar{b}_i$ is nondecreasing in $b_i$. Hence, $x_i < x_j$ implies $b_i < b_j$ and therefore $r_i(u) \le r_j(u)$ by Proposition F.2; pairs with $x_i = x_j$ contribute zero. Every summand is thus nonnegative. □

Because the position-wise sparse weights defined in Sec. F.3 normalize by routed mass, a larger $W_u^{\text{route}}(\gamma)$ does not directly scale an aggregate update. With fixed batching, allocation instead changes local step counts, which enter the routed computation scale $\psi_u$ and effective client mixture $\varpi_i(u)$ defined in that section.

**Remark F.1 (Scope of the monotonicity result)** *Proposition F.4 shows that increasing $\gamma$ assigns more idealized data mass to clients whose routed sub-supernets contain position $u$. It does not imply that the activation-weighted mass $\sum_{i=1}^{N} q_i(\gamma) h_i(u)$ is nondecreasing, because $h_i(u)$ need not be nondecreasing in client budget. Higher-budget clients have larger affordable sets and therefore different random-intermediate sampling pools. Activation monotonicity would require an additional assumption on the fixed cache and intermediate-variant sampling rule.*

### F.3 Position-wise sparse aggregation

Let $\Delta_{i,t}(u)$ be client $i$'s returned change at global position $u$. When $M_t^{\text{route}}(u) > 0$, define

$$\omega_{i,t}(u) = \frac{\mathbb{I}\{i \in \mathcal{S}_t\} n_i r_i(u)}{M_t^{\text{route}}(u)}, \tag{49}$$

and extend $\omega_{i,t}(u) = 0$ when $M_t^{\text{route}}(u) = 0$. The exact server update is

$$\boldsymbol{\theta}^{t+1}[u] = \begin{cases} \boldsymbol{\theta}^t[u] + \sum_{i=1}^{N} \omega_{i,t}(u)\Delta_{i,t}(u), & M_t^{\text{route}}(u) > 0, \\ \boldsymbol{\theta}^t[u], & M_t^{\text{route}}(u) = 0. \end{cases} \tag{50}$$

For a covered position, $\omega_{i,t}(u) \ge 0$ and $\sum_i \omega_{i,t}(u) = 1$. The effective client mixture is therefore position-dependent. Moreover, the denominator depends on $r_i(u)$ rather than $a_{i,\tau}(u)$: a routed client remains part of the position-wise average even when no sampled variant activates the coordinate. Such a return may be unchanged or may contain tensor-level optimizer effects. This routed-versus-activated distinction is the source of the aggregation and algorithmic perturbations represented below.

To connect this geometry to optimization, let $\mathcal{P}_i$ be the distribution of mixed-label tuples $\xi$ generated from client $i$'s data, and write $\ell_i^{\text{sup}}(\mathbf{a}) = \mathcal{L}_{\text{sup}}(z_i(\mathbf{a}; \boldsymbol{\theta}, \xi))$, with the dependence on $(\boldsymbol{\theta}, \xi)$ suppressed inside the expectation. The supervised reference loss is

$$f_i^{\text{sup}}(\boldsymbol{\theta}) = \mathbb{E}_{\xi \sim \mathcal{P}_i} \mathbb{E}_{\mathcal{V} \sim \nu_i} \Bigg[ \ell_i^{\text{sup}}(\mathbf{a}_i^{\max}) + (1 - \rho_{\text{kd}}) \sum_{\mathbf{a} \in \mathcal{V} \setminus \{\mathbf{a}_i^{\max}\}} \ell_i^{\text{sup}}(\mathbf{a}) \Bigg]. \tag{51}$$

The inner sum matches client gradient accumulation, so repeated activation of $u$ is already represented in $\partial_u f_i^{\text{sup}}$. With $p_i = n_i / N_{\text{train}}$ and $\sum_i n_i = N_{\text{train}}$, define

$$F_{\text{sup}}(\boldsymbol{\theta}) = \sum_{i=1}^{N} p_i f_i^{\text{sup}}(\boldsymbol{\theta}). \tag{52}$$

Distillation is excluded from this reference objective and enters the algorithmic discrepancy. Let $\mathcal{F}_t$ be the pre-round history, with $\mathbb{E}_t[\cdot] = \mathbb{E}[\cdot \mid \mathcal{F}_t]$. For $E$ local epochs and mini-batch size $B_{\text{mb}}$, let

$$B_i^{\text{loc}} = E \left\lceil \frac{n_i}{B_{\text{mb}}} \right\rceil \tag{53}$$

denote client $i$'s number of local optimizer steps per selected round. Canonically, $E = 1$ and $B_{\text{mb}} = 64$.

**Proposition F.5 (Auxiliary routed computation scale)** *Using the local-step counts in Eq. (53), define*

$$\beta_i(u) = \mathbb{E}_t[\omega_{i,t}(u)], \qquad \pi_u = \Pr[M_t^{\text{route}}(u) > 0]. \tag{54}$$

*Then $\sum_i \beta_i(u) = \pi_u$. Define*

$$\psi_u = \sum_{i=1}^{N} \beta_i(u) B_i^{\text{loc}}, \qquad \varpi_i(u) = \frac{\beta_i(u) B_i^{\text{loc}}}{\psi_u}, \tag{55}$$

*Here $\beta_i(u)$ is client $i$'s expected aggregate share and $\varpi_i(u)$ is the local-step-weighted effective client mixture. Consider the auxiliary frozen-gradient operator in which client $i$ returns $-\eta_t B_i^{\text{loc}} \nabla f_i^{\text{sup}}(\boldsymbol{\theta}^t)$, with no local drift, distillation, clipping, momentum, or weight decay. Its expected coordinate update satisfies*

$$\mathbb{E}_t[\boldsymbol{\theta}^{t+1}[u] - \boldsymbol{\theta}^t[u]] = -\eta_t \left[\psi_u \, \partial_u F_{\text{sup}}(\boldsymbol{\theta}^t) + e_t^{\text{agg}}(u)\right], \tag{56}$$

*where*

$$e_t^{\text{agg}}(u) = \psi_u \sum_{i=1}^{N} \big(\varpi_i(u) - p_i\big) \partial_u f_i^{\text{sup}}(\boldsymbol{\theta}^t). \tag{57}$$

*Moreover, if $B_{\min}^{\text{loc}} = \min_i B_i^{\text{loc}}$ and $B_{\max}^{\text{loc}} = \max_i B_i^{\text{loc}}$, then*

$$B_{\min}^{\text{loc}} \pi_u \le \psi_u \le B_{\max}^{\text{loc}} \pi_u. \tag{58}$$

**Proof.** For every cohort, $\sum_i \omega_{i,t}(u)$ equals one when $u$ is covered and zero otherwise, so taking expectation gives $\sum_i \beta_i(u) = \pi_u$. Substituting the auxiliary client return into Eq. (50) and taking conditional expectation gives

$$\begin{aligned} &-\eta_t \sum_{i=1}^{N} \beta_i(u) B_i^{\mathrm{loc}} \partial_u f_i^{\mathrm{sup}}(\boldsymbol{\theta}^t) \\ &\quad = -\eta_t \psi_u \sum_{i=1}^{N} \varpi_i(u) \partial_u f_i^{\mathrm{sup}}(\boldsymbol{\theta}^t). \end{aligned} \tag{59}$$

Adding and subtracting $\psi_u \sum_i p_i \partial_u f_i^{\mathrm{sup}}$ proves Eqs. (56) and (57). The bounds in Eq. (58) follow from $\sum_i \beta_i(u) = \pi_u$ and $B_{\min}^{\mathrm{loc}} \le B_i^{\mathrm{loc}} \le B_{\max}^{\mathrm{loc}}$. □

Proposition F.5 motivates, but does not prove, the functional form assumed for full FEAST below; A2–A4 collect the departures in discrepancy and noise terms. Activation is encoded in $\partial_u f_i^{\mathrm{sup}}$, whereas $\psi_u$ captures routed coverage, local steps, and sparse normalization.

## F.4 Routed-scale finite-horizon supervised stationarity

The result uses the following conditional decomposition and assumptions. All quantities and norms below are restricted to $\mathcal{U}^{\mathrm{srv}}$; coordinates outside it are neither updated by Eq. (50) nor used by served variants.

**A1: Smooth reference objective.** On the served support $\mathcal{U}^{\mathrm{srv}}$, which contains every coordinate updated by Eq. (50), $F_{\mathrm{sup}}$ is lower bounded by $F_{\mathrm{inf}}$ and has an $L_F$-Lipschitz gradient on the analyzed domain.

**A2: Conditional decomposition relative to the auxiliary scale.** Let $\boldsymbol{\Psi} = \mathrm{diag}(\psi_u)$, with $\psi_u$ defined by Eq. (55), and let $\boldsymbol{\Delta}_t = \boldsymbol{\theta}^{t+1} - \boldsymbol{\theta}^t$. Define

$$\mathbf{e}_t = -\frac{1}{\eta_t} \mathbb{E}_t[\boldsymbol{\Delta}_t] - \boldsymbol{\Psi} \nabla F_{\mathrm{sup}}(\boldsymbol{\theta}^t), \tag{60}$$

$$\boldsymbol{\zeta}_t = -\frac{1}{\eta_t} \boldsymbol{\Delta}_t - \boldsymbol{\Psi} \nabla F_{\mathrm{sup}}(\boldsymbol{\theta}^t) - \mathbf{e}_t. \tag{61}$$

Then $\mathbf{e}_t$ is $\mathcal{F}_t$-measurable, $\mathbb{E}_t[\boldsymbol{\zeta}_t] = \mathbf{0}$, and

$$\boldsymbol{\theta}^{t+1} = \boldsymbol{\theta}^t - \eta_t \left[ \boldsymbol{\Psi} \nabla F_{\mathrm{sup}}(\boldsymbol{\theta}^t) + \mathbf{e}_t + \boldsymbol{\zeta}_t \right]. \tag{62}$$

The operator $\boldsymbol{\Psi}$ is deterministic conditional on the realized partition and routed envelopes. Define

$$\psi_{\min} = \min_{u \in \mathcal{U}^{\mathrm{srv}}} \psi_u, \qquad \psi_{\max} = \max_{u \in \mathcal{U}^{\mathrm{srv}}} \psi_u. \tag{63}$$

Because every served position has positive routed coverage and $B_i^{\mathrm{loc}} \ge 1$, Eq. (58) implies

$$0 < \psi_{\min} \le \psi_u \le \psi_{\max} < \infty. \tag{64}$$

The activation profile is already encoded in the client supervised gradients; $\boldsymbol{\Psi}$ represents routed cohort coverage, local step counts, and position-wise sparse normalization.

**A3: Bounded update discrepancy.** Write $\mathbf{e}_t = \mathbf{e}_t^{\mathrm{agg}} + \mathbf{e}_t^{\mathrm{alg}}$. Almost surely, for constants $\varepsilon_{\mathrm{agg}}$ and $\varepsilon_{\mathrm{alg}}$,

$$\|\mathbf{e}_t\|_2 \le \varepsilon_{\mathrm{agg}} + \varepsilon_{\mathrm{alg}} =: \varepsilon. \tag{65}$$

The first label covers the position-dependent client mixture in Eq. (57); the second covers local drift, distillation, clipping, momentum, weight decay, and routed-but-unactivated optimizer updates. Only their sum is bounded here; no separate numerical estimate or small-error claim is made. The theorem studies dependence on $\psi_{\min}$ at fixed $\varepsilon$; the absolute discrepancy bound is conservative because some components may contract with gradient magnitude.

**A4: Bounded round noise.** Conditioned on $\mathcal{F}_t$,

$$\mathbb{E}_t[\boldsymbol{\zeta}_t] = \mathbf{0}, \qquad \mathbb{E}_t \|\boldsymbol{\zeta}_t\|_2^2 \le \sigma_{\mathrm{round}}^2. \tag{66}$$

A4 assumes a uniform conditional second-moment bound for the combined cohort, data, augmentation, and variant-sampling noise.

**Theorem F.1 (Finite-horizon supervised stationarity)** *Let*

$$\mathcal{H}_T = \sum_{t=0}^{T-1} \eta_t, \qquad \mathcal{Q}_T = \sum_{t=0}^{T-1} \eta_t^2, \tag{67}$$

*with $\mathcal{H}_T > 0$. Under A1–A4, if*

$$0 < \eta_t \le \frac{1}{4 L_F \psi_{\max}} \tag{68}$$

*for every analyzed round, then*

$$\begin{aligned} \frac{1}{\mathcal{H}_T} \sum_{t=0}^{T-1} \eta_t \mathbb{E} \|\nabla F_{\mathrm{sup}}(\boldsymbol{\theta}^t)\|_2^2 \le{}& \frac{2\big(F_{\mathrm{sup}}(\boldsymbol{\theta}^0) - F_{\mathrm{inf}}\big)}{\psi_{\min} \mathcal{H}_T} \\ &+ \left( \frac{2}{\psi_{\min}^2} + \frac{2 L_F \mathcal{Q}_T}{\psi_{\min} \mathcal{H}_T} \right) \varepsilon^2 \\ &+ \frac{L_F \mathcal{Q}_T}{\psi_{\min} \mathcal{H}_T} \sigma_{\mathrm{round}}^2. \end{aligned} \tag{69}$$

**Proof.** Write $\mathbf{g}_t = \nabla F_{\mathrm{sup}}(\boldsymbol{\theta}^t)$. By smoothness, Eq. (62), and $\mathbb{E}_t[\boldsymbol{\zeta}_t] = \mathbf{0}$,

$$\begin{aligned} \mathbb{E}_t F_{\mathrm{sup}}(\boldsymbol{\theta}^{t+1}) \le{}& F_{\mathrm{sup}}(\boldsymbol{\theta}^t) - \eta_t \langle \mathbf{g}_t, \boldsymbol{\Psi} \mathbf{g}_t + \mathbf{e}_t \rangle \\ &+ \frac{L_F \eta_t^2}{2} \left( \|\boldsymbol{\Psi} \mathbf{g}_t + \mathbf{e}_t\|_2^2 + \sigma_{\mathrm{round}}^2 \right). \end{aligned} \tag{70}$$

Weighted Young's inequality gives

$$-\langle \mathbf{g}_t, \mathbf{e}_t \rangle \le \frac{1}{4} \|\boldsymbol{\Psi}^{1/2} \mathbf{g}_t\|_2^2 + \|\boldsymbol{\Psi}^{-1/2} \mathbf{e}_t\|_2^2, \tag{71}$$

and

$$\|\boldsymbol{\Psi} \mathbf{g}_t\|_2^2 \le \psi_{\max} \|\boldsymbol{\Psi}^{1/2} \mathbf{g}_t\|_2^2, \qquad \|\boldsymbol{\Psi}^{-1/2} \mathbf{e}_t\|_2^2 \le \frac{\varepsilon^2}{\psi_{\min}}. \tag{72}$$

Using $\|\boldsymbol{\Psi} \mathbf{g}_t + \mathbf{e}_t\|_2^2 \le 2 \|\boldsymbol{\Psi} \mathbf{g}_t\|_2^2 + 2\varepsilon^2$ and the step-size condition in Eq. (68), Eq. (70) becomes

$$\begin{aligned} \mathbb{E}_t F_{\mathrm{sup}}(\boldsymbol{\theta}^{t+1}) \le{}& F_{\mathrm{sup}}(\boldsymbol{\theta}^t) - \frac{\eta_t}{2} \|\boldsymbol{\Psi}^{1/2} \mathbf{g}_t\|_2^2 \\ &+ \left( \frac{\eta_t}{\psi_{\min}} + L_F \eta_t^2 \right) \varepsilon^2 + \frac{L_F \eta_t^2}{2} \sigma_{\mathrm{round}}^2. \end{aligned} \tag{73}$$

Finally, $\|\mathbf{\Psi}^{1/2}\mathbf{g}_t\|_2^2 \geq \psi_{\min}\|\mathbf{g}_t\|_2^2$. Taking total expectations, summing Eq. (73), using $F_{\sup} \geq F_{\inf}$, and dividing by $\psi_{\min}\mathcal{H}_T/2$ proves Eq. (69). □

The theorem concerns the trained shared tensor before evaluation-time BN recalibration, not exact convergence of the complete KD objective or the recalibrated operating points. The quantity $\psi_{\min}$ is the minimum auxiliary routed computation scale; by Eq. (58), lower routed coverage can reduce this scale, and every term in the bound worsens as $\psi_{\min}$ decreases. We do not claim that $\varepsilon$ is small at the canonical $\rho_{\mathrm{kd}} = 0.5$; the theorem exposes how the routed scale, aggregate update discrepancy, and round noise enter the finite-horizon neighborhood rather than certifying a tight numerical radius.

### F.5 Canonical interpretation and empirical connection

For the canonical CIFAR-100 partition, $36 \leq n_i \leq 1906$, giving 1–30 local steps per client. Direct enumeration from the cache and budget vector gives $R_{\min} = 11$; with $N = 100$ and $m = 10$, the least-reachable profile is routed with probability

$$1 - \frac{\binom{89}{10}}{\binom{100}{10}} \approx 0.7062. \tag{74}$$

Thus, it is covered in approximately $70.62\%$ of uniformly sampled rounds. The accompanying reproduction script computes $\pi_u$ and $\psi_u$ for every routed profile under Proposition F.5, but these values do not numerically establish the unknown smoothness, discrepancy, or noise constants in Theorem F.1.

The structural results give a direct interpretation of the main-paper ablation. High-cost-exclusive positions are routed by a smaller upper tail of the client population. Proposition F.4 shows that increasing $\gamma$ shifts more idealized data mass toward that eligible tail, while Corollary F.1 shows that the activation-weighted mass also depends on FEAST's $h_i(u)$. These relationships are consistent with the CIFAR-100 largest-budget-minus-smallest-budget operating-point accuracy gap changing from $-4.29$ points at $\gamma = 0$ to $+6.17$ at $\gamma = 1$. The coupled protocol is not a volume-only intervention, and the theory does not predict this numerical gap.



\end{document}